\RequirePackage[svgnames]{xcolor}

\documentclass[11pt,a4paper]{mystyle}

\usepackage[all]{hypcap}
\usepackage[svgnames]{xcolor}
\usepackage[comma,authoryear,compress]{natbib}
\renewcommand{\cite}{\citep}

\usepackage{algorithm}
\usepackage{algorithmicx}
\usepackage{algpseudocode}
\usepackage{microtype}
\usepackage{booktabs}
\usepackage{float}
\usepackage{bigstrut}
\usepackage{multirow}

\usepackage{amsmath}
\usepackage{amssymb}
\usepackage{mathtools}
\usepackage{amsthm}
\usepackage{mathrsfs}
\usepackage{dsfont}
\usepackage{enumitem}
\usepackage{graphicx}
\usepackage{cleveref}
\usepackage{bxcoloremoji}

\usepackage{float}

\usepackage[utf8]{inputenc} %
\usepackage[T1]{fontenc}    %
\usepackage{hyperref}       %
\usepackage{url}            %
\usepackage{booktabs}       %
\usepackage{amsfonts}       %
\usepackage{nicefrac}       %
\usepackage{microtype}      %
\usepackage{amssymb}
\usepackage{wrapfig}
\usepackage{lipsum}
\usepackage{enumitem}
\usepackage{stackengine}
\usepackage[font=small,labelfont=bf]{caption}
\usepackage{color}
\usepackage{adjustbox}
\usepackage{soul}
\usepackage{rotating}
\usepackage{makecell}

\usepackage{xspace}
\usepackage{dsfont}
\usepackage{pifont}

\newcommand{\Tref}[1]{Table~\ref{#1}}

\newcommand{\Fref}[1]{Figure~\ref{#1}}

\newcommand{\Sref}[1]{\S~\ref{#1}}
\newcommand{\Apdxref}[1]{Appendix~\ref{#1}}

\makeatletter
\DeclareRobustCommand\onedot{\futurelet\@let@token\@onedot}
\def\@onedot{\ifx\@let@token.\else.\null\fi\xspace}
\def\eg{\emph{e.g}\onedot}

\def\ie{\emph{i.e}\onedot}

\makeatother


\usepackage{soul}
\definecolor{myblue}{RGB}{218,227,243}
\definecolor{mypink}{RGB}{251,229,214}
\definecolor{myred}{RGB}{252,150,148}
\definecolor{mygreen}{RGB}{226,240,217}

\definecolor{myyellow}{HTML}{FFE083}
\definecolor{mypurple}{HTML}{CD93E7}
\definecolor{mygreen1}{HTML}{AFD5AB}
\definecolor{myblue1}{HTML}{98C9F5}
\definecolor{myred1}{HTML}{DE9B99}

\title{\textit{Logical Consistency as a Bridge:} 
Improving LLM Hallucination Detection via Label Constraint Modeling between Responses and Self-Judgments}

\runningtitle{\textit{Logical Consistency as a Bridge:} 
Improving LLM Hallucination Detection via Label Constraint Modeling between Responses and Self-Judgments}

\author{\href{https://summerrice.github.io/}{\textcolor{black}{Hao Mi}}}
\author{\href{https://sheng-qiang.github.io/}{\textcolor{black}{Qiang Sheng}}}
\author{Shaofei Wang}
\author{\href{https://beanandrew.github.io/}{\textcolor{black}{Beizhe Hu}}}
\author{\href{https://scholar.google.com/citations?user=sv7uxi4AAAAJ}{\textcolor{black}{Yifan Sun}}}
\author{\href{https://zhengjiawa.github.io/}{\textcolor{black}{Zhengjia Wang}}}
\author{Hengqi Zeng}
\author{\href{https://www.yang-li.cn/}{\textcolor{black}{Yang Li}}}
\author{\href{https://scholar.google.com/citations?user=hGZwK0cAAAAJ&hl=en}{\textcolor{black}{Danding Wang}}}
\author{\href{https://scholar.google.com/citations?user=fSBdNg0AAAAJ&hl=zh-CN&oi=ao}{\textcolor{black}{Juan Cao}}}

\affil{Institute of Computing Technology, Chinese Academy of Sciences}
\affil{University of Chinese Academy of Sciences}

\emails{{\{\href{mailto:mihao24s@ict.ac.cn}{mihao24s},\href{mailto:shengqiang18z@ict.ac.cn}{shengqiang18z}\}@ict.ac.cn}}

\begin{document}

\begin{abstract}
Large Language Models (LLMs) are prone to factual hallucinations, risking their reliability in real-world applications.
Existing hallucination detectors mainly extract micro-level intrinsic patterns for uncertainty quantification or elicit macro-level self-judgments through verbalized prompts.
However, these methods address only a single facet of the hallucination, focusing either on implicit neural uncertainty or explicit symbolic reasoning, thereby treating these inherently coupled behaviors in isolation and failing to exploit their interdependence for a holistic view.
In this paper, we propose \textbf{LaaB} (\textbf{L}ogical Consistency-\textbf{a}s-\textbf{a}-\textbf{B}ridge), a framework that bridges neural features and symbolic judgments for hallucination detection.
LaaB introduces a ``meta-judgment'' process to map symbolic labels back into the feature space. By leveraging the inherent logical bridge where response and meta-judgment labels are either the same or opposite based on the self-judgment's semantics, LaaB aligns and integrates dual-view signals via mutual learning and enhances the hallucination detection.
Extensive experiments on 4 public datasets, across 4 LLMs, against 8 baselines demonstrate the superiority of LaaB.

\vspace{3mm}

\coloremojicode{1F4C5} \textbf{Date}: May 6th, 2026

\coloremojicode{1F3E0} \textbf{Project}: \href{https://summerrice.github.io/LaaB/}{https://summerrice.github.io/LaaB}

\coloremojicode{1F4AC} \textbf{Venue}: ACL 2026

\end{abstract}

\maketitle
\vspace{0mm}

\section{Introduction}

Large language models (LLMs) have shown impressive capabilities across diverse tasks \citep{wei2022h, hu2024bad, zhao2026agentic, chen2026simulating}. 
However, their reliability in real-world applications is compromised by factual hallucinations: outputs that appear plausible but contradict verified facts or commonsense, even without malicious prompting \citep{huang2025survey}. 
As recent studies indicate that hallucinations may be an inherent property of LLMs rather than a fully solvable error, their complete elimination remains elusive \citep{xu2024hallucination,mohsin2025fundamental}. 
Consequently, accurate hallucination detection is a critical requirement for maintaining reliable and trustworthy LLM-based systems \citep{ji2023survey, Liu24tutorials, zhang2025siren, hu2025llm}.

\begin{wrapfigure}{r}{0.45\textwidth}
  \centering
  \vspace{-1em}
  \includegraphics[width=0.43\textwidth]{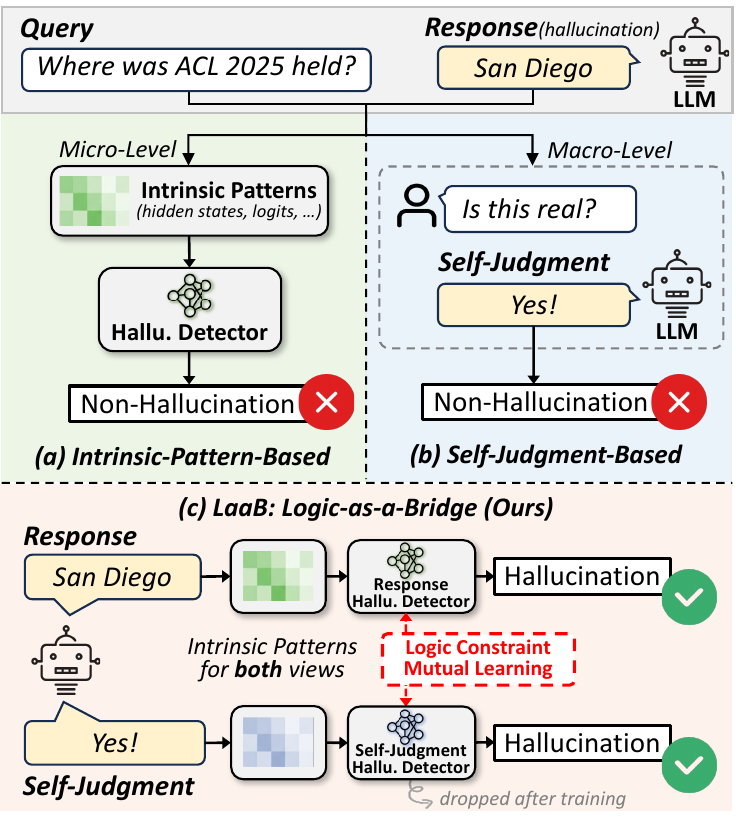}
  \caption{Comparison between existing hallucination detection methods (a-b) and our proposed LaaB (c). Unlike methods that rely solely on micro-level intrinsic patterns (a) or macro-level symbolic judgment (b), LaaB bridges these two views by enforcing logical consistency via logic-constraint mutual learning.}
  \label{fig:1}
  \vspace{-1em}
\end{wrapfigure}

Hallucination detection is generally formulated as a binary classification task that predicts the factuality of LLM responses. Besides applying fact-checking that relies on reliable external knowledge~\cite{min2023factscore}, recent work looks \textit{inside} LLMs by 1) extracting intrinsic patterns, or 2) using LLMs themselves as judges. 
The intrinsic-pattern-based detectors exploit the LLM's behavioral patterns during generation to quantify its uncertainty of the response (see \Fref{fig:1}(a)), including generation consistency~\citep{manakul2023selfcheckgpt, farquhar2024detecting}, output confidence~\cite{guo2017calibration}, hidden states~\cite{azaria2023internal}, and attention maps~\cite{chuang2024lookback}, etc. 
These methods capture the nuanced internal signals from a micro perspective, but these metrics may lack proper calibration, resulting in the failure to identify high-certainty hallucinations~\citep{tan2025too,wen2024mitigating,zhou2024relying,simhi2025trust}.
In contrast, the detectors based on self-judgments directly obtain LLMs' verbal judgments through factuality-oriented prompting (see \Fref{fig:1}(b)), assuming that LLMs may elicit different knowledge when switching the role from answering to judging~\citep{ji2023towards, li2025hallucination}.
This paradigm exploits the macro-level judgment, but the verbal judgment suffers from self-preference bias \citep{wataoka2024self,panickssery2024llm} or overthinking issues~\citep{zhang2025understanding, su2025between}, possibly leading to ``secondary'' hallucination.
This motivates us to explore: \textbf{How to effectively integrate the micro-level intrinsic signals and macro-level self-judgments for more accurate hallucination detection?}

To perform effective integration, we propose a hallucination detection method named \textbf{LaaB} (\textbf{L}ogical Consistency-\textbf{a}s-\textbf{a}-\textbf{B}ridge; see \Fref{fig:1}(c)).
The key challenge in the integration is to build a joint, learnable framework for both the neural features derived from intrinsic patterns and the symbolic judgments from the LLM itself.
To address this issue, LaaB exploits the inherent logical constraint between the response and self-judgment as a bridge.
First, we map the symbolic self-judgment back to the feature space to make it possible to optimize the hallucination detection via joint learning.
Our idea starts from a simple but crucial fact: The LLM's self-judgment on its response is \textit{also} a response that may be hallucinatory, and the intrinsic signal of the judgment generation may reveal its own veracity.
By applying intrinsic-pattern-based methods on the self-judgment (\ie, meta-judgment), LaaB obtains the learned features from the quantification of the self-judgment uncertainty.
Subsequently, we transform the hallucination prediction of the self-judgment into that of the original response by leveraging the inherent logic constraint: the response and the self-judgment would share the same factuality label if the self-judgment claims that the response is truthful; otherwise, their labels should be opposite.
Based on this logic rule, LaaB can obtain two aligned predictions for the original response from two neural modules of different views,  thus enhancing the final judgment of hallucination.
A mutual learning strategy is finally adopted for the whole optimization.

Our contributions are summarized as follows:
\vspace{-1mm}
\begin{itemize}[leftmargin=*]
    \item \textbf{Concept:} We propose to see LLMs' self-judgment as a special response that can be checked by another hallucination detector, whose results will bridge the logic constraint that enables the integration of the predictions from both the response and self-judgment views.
    \vspace{1mm}
    \item \textbf{Method:} We design LaaB, which bridges the prediction signals from both intrinsic patterns and self-judgments, and builds a mutual learning framework for accurate hallucination detection.
    \vspace{1mm}
    \item \textbf{Performance:} Experiments on 4 public datasets, across 4 LLMs, against 8 baselines show that LaaB can effectively enhance the performance in hallucination detection without introducing significant additional inference cost.
\end{itemize}

\section{Related Work}

\subsection{Hallucination Detection}
Hallucination detection aims to evaluate the factuality of LLM outputs. 
Given that conventional fact-checking relies heavily on external knowledge retrieval and evidence verification~\citep{hu-etal-2024-knowledge, min2023factscore, wang2024factcheck, wan2025fastfact, chernfactool}, we focus on detection methods that leverage the LLM's internal signals, which fall into two categories: Intrinsic-pattern-based methods and self-judgment-based methods.

\noindent\underline{\textbf{Intrinsic-pattern-based methods}} leverage the signals generated during the inference, positing that LLMs exhibit distinct internal behaviors when hallucinating compared to when generating factual content, typically including hidden states, prediction logits, and attention scores.
\textit{Hidden-state-based methods} assume that the truthful and hallucinated boundaries are encoded in the LLM's latent space, generally training classifiers on layer-wise hidden states~\citep{li2023inferencetime, azaria2023internal, su2024unsupervised, du2024haloscope, liu2024universal, park2025steer, ni-etal-2025-towards} or activation dynamics~\citep{wanglatent, zhang2025icr}.
\textit{Logit-based methods} interpret the output probability distribution as a proxy for model confidence, where lower confidence or higher entropy often correlates with hallucination~\citep{jiang2024large, ma2025estimating, vashurin2025uncertainty, vazhentsev2025unconditional, li-etal-2025-towards, tan2026basecal}.
\textit{Attention-based methods} exploit the information flow, assuming hallucinations stem from improper attention to the context, and identify anomalies through attention map distributions~\citep{chuang2024lookback, liu2025attention, binkowski2025hallucination, qi2026detectingcontextual}.
Despite capturing micro-level uncertainty, they often lack semantic calibration, failing to identify hallucinations with high confidence~\citep{tan2025too, simhi2025trust, li2025conftuner}.

\noindent\underline{\textbf{Self-judgment-based methods}} leverage the semantic reasoning capabilities of LLMs to assess factuality through verbal interaction.
Early works investigated the self-evaluation feasibility~\citep{yin2023large, xiongcan}, while subsequent research introduced mechanisms like self-correction~\citep{ji2023towards, dhuliawala-etal-2024-chain, yuanyige, metaselfcorrecting} and multi-agent debating~\citep{liu2025long, sun2025towards} to elicit accurate judgments.
However, it inherently relies on the model's generation capabilities, making it susceptible to self-preference bias and ``evaluative hallucination''~\citep{wataoka2024self, zhang2025understanding}.
To address these limitations, we argue that intrinsic signals and verbal judgments are complementary: the former quantifies micro-level uncertainty, while the latter provides logical anchoring.
In this work, we propose to bridge these two perspectives by viewing the self-judgment not as a ground truth, but as another generative behavior subject to hallucination. 
By modeling the logical constraint between the \textit{response} and the \textit{self-judgment}, we integrate them into a mutual learning framework to enhance detection accuracy.

\subsection{Mutual Learning}
Mutual learning is a paradigm in which peer networks learn from each other to improve performance and generalization. 
\citet{zhang2018deep} introduced deep mutual learning, where multiple students are jointly trained to align their output distributions with peers. Subsequent works regard mutual learning as a process of online knowledge distillation where the peer predictions serve as a dynamic teacher, and develop variants by introducing ensemble learning~\citep{guo2020online,tan2022online} and contrastive learning~\citep{yang2023online}. 
In this paper, we adopt mutual learning to bridge the individual learning of hallucination detection modules for the original response and self-judgment (which serve as the peers in our method).

\section{Task Formulation}
\label{sec:form}
We formulate hallucination detection as a binary classification task. Given a user query $Q_r$ and an LLM-generated response $O_r$, the primary goal is to determine the factuality label $L_r \in \{0, 1\}$, where $L_r=1$ denotes a factual response and $L_r=0$ denotes a hallucination response.

For an intrinsic-pattern-based detector $D_r$, it maps intrinsic model features $F_r$ (derived during the generation process of $O_r$) to a predicted probability distribution $S_r$ over the labels, and then gets the predicted label $\hat{L}_r$. 
For a self-judgment-based detector, it generates a verbal judgment $O_j \in \{\text{``Yes''}, \text{``No''}\}$ based on the original response $O_r$ and a factuality evaluation prompt $Q_j$, where ``Yes'' and ``No'' can also correspond to the predicted label $\hat{L}_r=1$ and $0$, respectively.
For the meta-judgment we will detail in \Sref{sec:method}, similar to $D_r$, a secondary detector $D_j$ maps the intrinsic features $F_j$ for $O_j$ to the ground-truth label of the judgment factuality $L_j \in \{0, 1\}$. 
The consolidated list of notations used in the LaaB framework is presented in Table \ref{tab:notation}.

\begin{table}[t]
\centering
\renewcommand{\arraystretch}{1.25}
\captionsetup{justification=centering}
\caption{Summary of notations in LaaB Framework}
\begin{tabularx}{\textwidth}{l X @{\hspace{6pt}} !{\vrule width 0.6pt} @{\hspace{6pt}} l X}
\toprule
\multicolumn{2}{c}{\textbf{\textit{Response}}} & \multicolumn{2}{c}{\textbf{\textit{Self-Judgment}}} \\
\midrule
$Q_r$ & User query & $Q_j$ & Evaluation prompt on $O_r$\\
$O_r$ & LLM response to $Q_r$ & $O_j \in \{\text{``Yes'', ``No''}\}$ & LLM response to $Q_j$ (Verbal judge)\\
$F_r$ & Intrinsic features for $O_r$ & $F_j$ & Intrinsic features for $O_j$ \\
$D_r$ & Detector over $F_r$ & $D_j$ & Detector over $F_j$ \\
$L_r \in \{0,1\}$ & Ground-truth for $O_r$ (1: factual) & $L_j \in \{0,1\}$ & Ground-truth for $O_j$ (Meta judge) \\
$\hat{L}_r \in \{0,1\}$ & Predicted label for $O_r$ & $\hat{L}_j \in \{0,1\}$ & Predicted label for $O_j$\\
$S_r$ & Predicted distribution from $D_r$ & $S_j$ & Predicted distribution from $D_j$ \\
\bottomrule
\end{tabularx}
\label{tab:notation}
\end{table}

\section{Proposed Method: LaaB}
\label{sec:method}

\begin{figure}[htbp]
\centering
\includegraphics[width=0.95\textwidth]{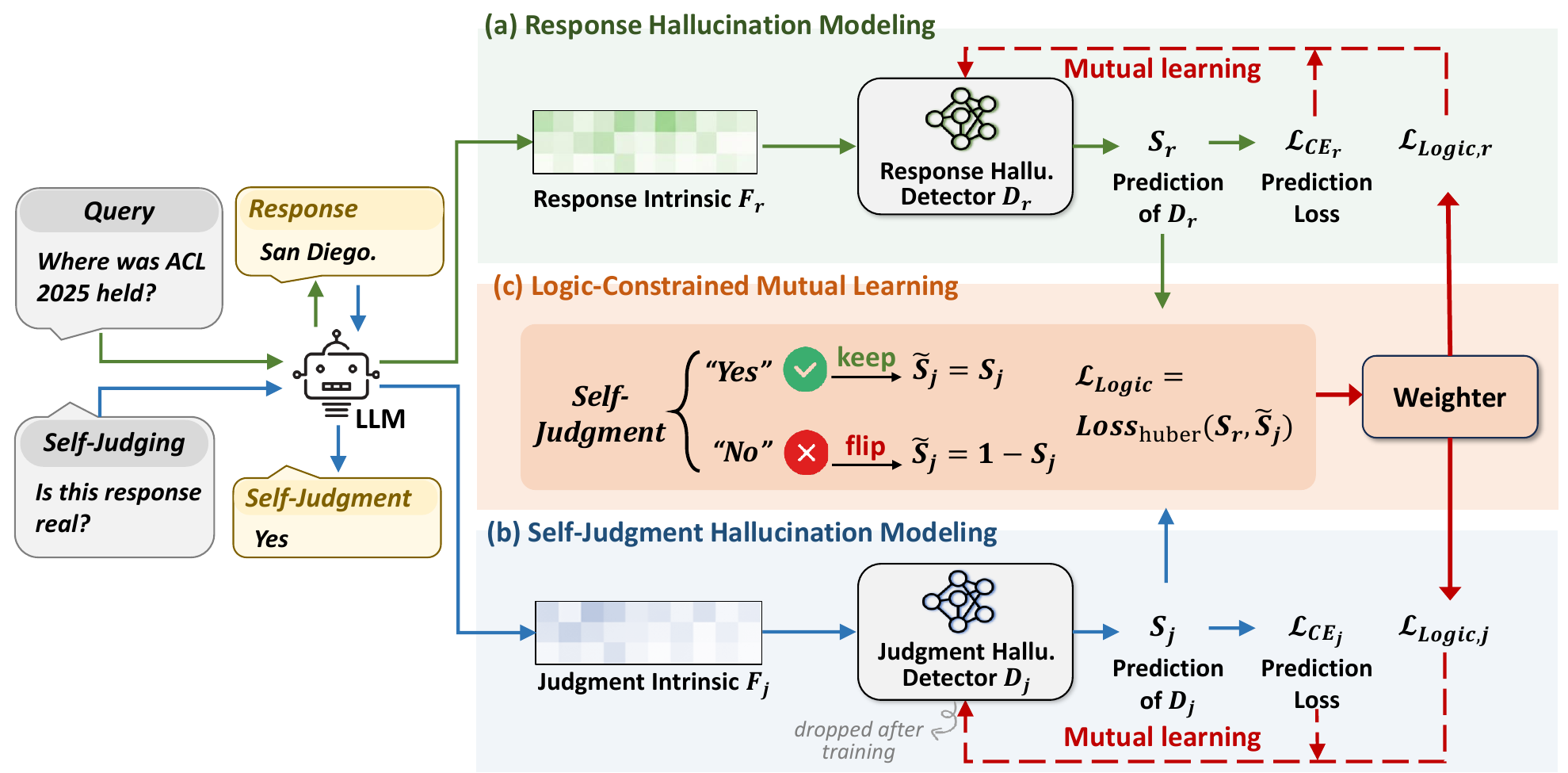}
\caption{Overall architecture of \textbf{LaaB}. Given a user query and corresponding response, LaaB first performs \textbf{(a)} Response Hallucination Modeling, extracting intrinsic features from the response generation to capture implicit uncertainty. \textbf{(b)} Self-Judgment Hallucination Modeling introduces a meta-judgment process that analyzes the elicited verbal judgment to mitigate evaluative biases. Finally, \textbf{(c)} Logic-Constrained Mutual Learning bridges these dual views by enforcing logical consistency between response and judgment predictions, utilizing their inherent dependency for robust joint optimization.}
\label{fig:framework}
\end{figure}

\Fref{fig:framework} overviews our proposed LaaB, which first models response hallucination and self-judgment hallucination, respectively, and then adopts logic-constrained mutual learning to guide the joint optimization. We detail the components' design below.

\subsection{Response Hallucination Modeling}
\label{sec:intrinsic method}

This module detects hallucinations using intrinsic patterns for generating $O_r$. We utilize an MLP-based detector $D_r$ that accepts features $F_r$ and outputs a probability distribution $S_r = (S_{r\_hallu}, S_{r\_real})$. The extracted features include:

\noindent\underline{\textbf{Hidden States ($H_r$).}} 
Following \citet{azaria2023internal}, we assume the hidden representation of final token in a sequence aggregates semantic information. We feed the concatenated sequence pair $(Q_r, O_r)$ into the LLM and extract the hidden state of the last token at the validation-optimal layer $K_{\text{val},r}$, denoted as $H_r$.

\noindent\underline{\textbf{Prediction Logits ($P_r$).}} We adopt the Logits Lens hypothesis \citep{logit-lens,jiang2024large}, which assumes that all hidden layers of the LLM share the same unembedding space with the output layer.  Let $P_{ll} $ represent the layer-wise probabilities of the generated tokens in $O_r$. We aggregate $P_{ll}$ via a single Transformer layer followed by mean-pooling to obtain the sequence-level feature $P_r$.

\noindent\underline{\textbf{Attention Scores ($A_r$).}} 
Following \citet{chuang2024lookback}, we utilize the ``Lookback'' ratio, which quantifies the token-level attention assigned to preceding context components. Building upon this idea, we further adapt it to our context for factual hallucination detection. For each token in $O_r$, we compute attention ratios targeting four segments: the system prompt, query $Q_r$, response trigger, and preceding tokens of $O_r$. We pool these token-wise ratios and select the top-$P$ informative heads based on KL divergence to form the feature $A_r$.

The details of feature extraction are provided in Appendix \ref{app:feature}. In the implementation, $D_r$ uses one feature in $\{H_r, P_r, A_r\}$ as the input $F_r$ and is trained via cross-entropy loss $\mathcal{L}_{\text{CE,r}}$ against $L_r$.

\subsection{Self-Judgment Hallucination Modeling}
\label{sec:judgment method}

We define \textit{evaluative hallucination} as the scenario where an LLM acting as a judge produces an incorrect assessment $O_j$. Rather than accepting $O_j$ as ground truth, we treat it as a generated artifact with a learnable factuality label $L_j$. We train a detector $D_j$ to predict $L_j$ based on intrinsic features $F_j$ extracted during the generation of $O_j$.
Similar to \S\ref{sec:intrinsic method}, we extract features $F_j \in \{H_j, P_j, A_j\}$, with specific adaptations for the judgment context:

\noindent\underline{\textbf{Hidden States ($H_j$).}} The states are extracted from the last token of the judgment $O_j$ at layer $K_{\text{val},j}$.

\noindent\underline{\textbf{Prediction Logits ($P_j$).}} Based on Logits Lens, we first extract the layer-wise probability distribution of the first token of $O_j$. Then we construct semantic sets $V_{\text{yes}}$ and $V_{\text{no}}$ (grouping synonyms for ``Yes''/``No'', Appendix \ref{app:logits}) and aggregate their probabilities into $P_{\text{yes}}$ and $P_{\text{no}}$. To emphasize the contrast, the input feature is constructed as:
    \begin{equation}
    P_j =
    \begin{cases} 
    P_{\text{yes}} \oplus (P_{\text{yes}} - P_{\text{no}}), & \text{if } O_j = \text{``Yes''}, \\[1ex]
    P_{\text{no}} \oplus (P_{\text{no}} - P_{\text{yes}}), & \text{if } O_j = \text{``No''}.
    \end{cases}
    \end{equation}
    
\noindent\underline{\textbf{Attention Scores ($A_j$).}}
We segment the judgment context $Q_j$ into six components (Framing, Query, Response, Eval\_Query, Format, and Trigger) and compute the attention ratios of the judgment token over these segments to form $A_j$.

\subsection{Logic-Constrained Mutual Learning}
\label{sec:mutual}

\noindent\underline{\textbf{Logical Dependency.}}
In the self-judgment setting, the verbal judgment $O_j$ implies the factuality prediction of the response $O_r$, and we can derive the logical consistency between $L_r$ and $L_j$. Specifically, when $O_j$ is “Yes”, the LLM predicts $\hat{L}_r = 1$. If $L_j = 1$, then the above prediction is correct and $L_r = 1$; otherwise, the prediction is incorrect and $L_r = 0$. Similarly, when $O_j$ is “No”, the relationship between the labels is reversed. Logical dependency is summarized in Table~\ref{tab:label_logic}.

\begin{table}[t]
    \centering
    \renewcommand{\arraystretch}{1.4}
    \setlength\tabcolsep{10pt}
    \caption{Logic constraints between the truth value of the response factuality and self-judgment factuality based on the LLM self-judgment.}
    \begin{tabularx}{\textwidth}{c X c}
    \toprule
    \textbf{Self-Judgment} ($O_j$) & \textbf{Interpretation} & \textbf{Target} ($L_r$) \\
    \midrule
    ``Yes'' & \textbf{Affirmation}: The LLM supports the original response. Keep labels consistent. & $L_j$ \\
    ``No''  & \textbf{Negation}: The LLM refutes. Reverse the factuality. & $1 - L_j$ \\
    \bottomrule
    \end{tabularx}
    \label{tab:label_logic}
\end{table}

\noindent\underline{\textbf{Framework.}}
Detectors $D_r$ and $D_j$ produce probability distributions $S_r = (S_{r, \text{hallu}}, S_{r, \text{real}})$ and $S_j = (S_{j, \text{hallu}}, S_{j, \text{real}})$, respectively. To enforce logical consistency, we employ the Huber loss~\cite{huberloss}, denoted as $\mathcal{L}_{\text{Huber}}$, to align the scalar probabilities of the two detectors. The logic-constrained loss $\mathcal{L}_{\text{Logic}}$ is defined as:
\begin{equation}
\mathcal{L}_{\text{Logic}}\! \!=\!\!
\begin{cases}
\mathcal{L}_{\text{Huber}}(S_{r, \text{hallu}}, S_{j, \text{hallu}}), \!\!\!& \text{if } O_j = \text{``Yes''}, \\
\mathcal{L}_{\text{Huber}}(S_{r, \text{hallu}}, S_{j, \text{real}}),\!\!\! & \text{if } O_j = \text{``No''}.
\end{cases}
\end{equation}
This loss encourages $D_r$ and $D_j$ to learn robust representations by aligning their predictions according to the logical dependency. More details about Huber loss are provided in Appendix \ref{app:huber}.

To prevent mutual degradation (where a weaker detector misleads a stronger one), we introduce a confidence-aware weighting mechanism. We assume a detector with higher confidence on the ground truth label possesses better representations. For a sample pair, we compute a weight based on the ratio of the peer's confidence to the self's confidence. The specific logic losses for $D_r$ and $D_j$ are:
\begin{align}
\mathcal{L}_{\text{Logic}, r} &= \log\left(1 + \frac{S_j(L_j)}{S_r(L_r)}\right) \cdot \mathcal{L}_{\text{Logic}}, \\
\mathcal{L}_{\text{Logic}, j} &= \log\left(1 + \frac{S_r(L_r)}{S_j(L_j)}\right) \cdot \mathcal{L}_{\text{Logic}},
\end{align}
where $S(L)$ denotes the predicted probability of the ground truth class.
Finally, the total training objective for each detector ($* \in \{r, j\}$) combines the cross-entropy loss and the weighted logic loss:
\begin{equation}
\mathcal{L}_* = \mathcal{L}_{\text{CE}, *} + \alpha_* \mathcal{L}_{\text{Logic}, *}.
\end{equation}
Here, $\alpha_*$ is a batch-level balancing coefficient dynamically computed using the ratio of gradient norms with respect to the last-layer parameters $\theta_*^{-1}$, ensuring stable optimization:
\begin{equation}
\alpha_* = \frac{\left\| \nabla_{\theta_*^{-1}} \mathcal{L}_{\text{CE}, *} \right\|_2}{\left\| \nabla_{\theta_*^{-1}} \mathcal{L}_{\text{Logic}, *} \right\|_2 + \epsilon}.
\end{equation}

\noindent\underline{\textbf{Training Strategy.}}
We adopt a two-stage training strategy. In the first stage, $D_r$ and $D_j$ are trained asynchronously in a round-robin manner; when one converges, it is frozen while the other continues. In the second stage, both detectors are jointly fine-tuned using the combined loss:
\begin{equation}
\mathcal{L}_{\text{Joint}} = \mathcal{L}_{\text{CE}, r} + \mathcal{L}_{\text{CE}, j} + \alpha \mathcal{L}_{\text{Logic}}.
\end{equation}

\noindent\underline{\textbf{Inference.}}
During inference, only $D_r$ is deployed. This allows the system to benefit from the knowledge distilled from $D_j$ via mutual learning, without incurring the additional computational costs associated with the judgment generation. The pseudocode for the LaaB training and inference procedures is provided in Appendix \ref{app:algorithm}.

\section{Experiments}

\subsection{Experimental Settings}
\noindent\textbf{Datasets.} 
We utilize four widely used datasets for factualness hallucination detection: TriviaQA, MMLU, NQ\_Open, and HaluEval (\Apdxref{app:dataset}). 
For each dataset, we 1) prompt the LLM to generate responses to the given query; 2) prompt the LLM to self-evaluate its responses using the template (\Apdxref{app:prompt}); 3) annotate the factuality of responses and judgments based on the ground-truth (\Apdxref{app:label}); and 4) split the resulting dataset with a 7:1:2 ratio for the training, validation, and testing sets.

\noindent\textbf{Large Language Models.}
We use four commonly-used open-source LLMs, including Llama-3.1-8B-Instruct, Llama-3.1-70B-Instruct, Qwen-2.5-32B-Instruct, and Mistral-7B-Instruct-v0.3, which cover different families and scales (Appendix \ref{subsec:llms}).

\noindent\textbf{Baselines.}
We include self-judgment-based detector Self-Judge~\citep{kadavath2022language} and five intrinsic-pattern-based detectors. The trainable detectors using internal model representations are detailed in \S\ref{sec:intrinsic method}, including SAPLMA (with hidden states), Logits Lens (with logits), and Lookback Lens (with attention patterns). The latter two include consistency-based methods, SelfCheckGPT~\citep{manakul2023selfcheckgpt} and EigenScore~\citep{chen2024inside}, which are detailed in Appendix \ref{app:baselines}.

\noindent\textbf{Evaluation Metrics.}
Macro F1 (macF1) and accuracy (Acc) are adopted as evaluation metrics for hallucination detection, reflecting class-level and instance-level performance, respectively.

\begin{table}[htbp]
  \centering\small
   \caption{Performance comparison of baselines and LaaB in hallucination detection. \textbf{Bolded} numbers denote that the use of LaaB is better-performing than its corresponding base version. \underline{Underlined} numbers are the highest in each column within each LLM group.}
   \setlength{\tabcolsep}{3.5pt} 
    \begin{tabular}{
      >{\centering\arraybackslash}m{1.9cm} 
      >{\raggedright\arraybackslash}m{2.5cm} 
      @{}c@{} 
      cc      
      >{\centering\arraybackslash}m{0.0cm} 
      cc      
      >{\centering\arraybackslash}m{0.0cm} 
      cc      
      >{\centering\arraybackslash}m{0.0cm} 
      cc      
      >{\centering\arraybackslash}m{0.0cm} 
      cc      
    }
    \toprule
     &  & \textbf{} & \multicolumn{2}{c}{\textbf{TriviaQA}} & \textbf{} & \multicolumn{2}{c}{\textbf{MMLU}} & \textbf{} & \multicolumn{2}{c}{\textbf{NQ\_Open}} & \textbf{} & \multicolumn{2}{c}{\textbf{HaluEval}} & \textbf{} & \multicolumn{2}{c}{\textbf{Average}} \\ \cline{4-5} \cline{7-8} \cline{10-11} \cline{13-14} \cline{16-17}
    \multirow{-2}{*}{\textbf{LLM}} & \multirow{-2}{*}{\textbf{Method}} &  & macF1 & Acc &  & macF1 & Acc &  & macF1 & Acc &  & macF1 & Acc & & macF1 & Acc \\ \midrule
    
     & Self-Judge &  & 67.67 & 71.36 &  & 57.29 & 65.13 &  & 59.12 & 59.14 &  & 60.89 & 62.78 & & 61.24 & 64.60 \\
     & SelfCheckGPT &  & 76.34 & 81.98 &  & 59.78 & 63.56 &  & 69.77 & 73.17 &  & 74.41 & 74.49 & & 70.08 & 73.30 \\
     & Eigen-Score &  & 69.92 & 75.58 &  & 57.88 & 63.47 &  & 64.82 & 67.07 &  & 66.34 & 66.81 & & 64.74 & 68.23 \\
     & SAPLMA &  & 77.05 & 79.87 &  & 69.96 & 70.75 &  & 73.01 & 75.15 &  & 75.70 & 75.88 & & 73.93 & 75.41 \\
     & {\hspace{15pt}\textit{+LaaB}} & \cellcolor[HTML]{ECF4FF} & \cellcolor[HTML]{ECF4FF}\underline{\textbf{78.74}} & \cellcolor[HTML]{ECF4FF}\underline{\textbf{82.09}} & \cellcolor[HTML]{ECF4FF} & \cellcolor[HTML]{ECF4FF}\textbf{71.77} & \cellcolor[HTML]{ECF4FF}\textbf{73.25} & \cellcolor[HTML]{ECF4FF} & \cellcolor[HTML]{ECF4FF}\underline{\textbf{73.10}} & \cellcolor[HTML]{ECF4FF}\underline{\textbf{77.13}} & \cellcolor[HTML]{ECF4FF} & \cellcolor[HTML]{ECF4FF}\underline{\textbf{77.20}} & \cellcolor[HTML]{ECF4FF}\underline{\textbf{77.60}} & \cellcolor[HTML]{ECF4FF} & \cellcolor[HTML]{ECF4FF}\underline{\textbf{75.20}} & \cellcolor[HTML]{ECF4FF}\underline{\textbf{77.52}} \\
     & Logits Lens &  & 72.50 & 74.81 &  & 65.27 & 66.07 &  & 62.63 & 63.26 &  & 72.21 & 72.48 & & 68.15 & 69.16 \\
     & {\hspace{15pt}\textit{+LaaB}} & \cellcolor[HTML]{ECF4FF} & \cellcolor[HTML]{ECF4FF}\textbf{75.11} & \cellcolor[HTML]{ECF4FF}\textbf{78.88} & \cellcolor[HTML]{ECF4FF} & \cellcolor[HTML]{ECF4FF}\textbf{66.72} & \cellcolor[HTML]{ECF4FF}\textbf{70.65} & \cellcolor[HTML]{ECF4FF} & \cellcolor[HTML]{ECF4FF}\textbf{65.72} & \cellcolor[HTML]{ECF4FF}\textbf{68.90} & \cellcolor[HTML]{ECF4FF} & \cellcolor[HTML]{ECF4FF}\textbf{72.29} & \cellcolor[HTML]{ECF4FF}\textbf{73.04} & \cellcolor[HTML]{ECF4FF} & \cellcolor[HTML]{ECF4FF}\textbf{69.96} & \cellcolor[HTML]{ECF4FF}\textbf{72.87} \\
     & Lookback Lens &  & 71.65 & 74.45 &  & \underline{72.74} & \underline{73.96} &  & 68.60 & 70.88 &  & 75.13 & 75.49 & & 72.03 & 73.70 \\
    \multirow{-9}{*}{\makecell[l]{\textbf{Llama-3.1-}\\\textbf{8B-Instruct}}} & {\hspace{15pt}\textit{+LaaB}} & \cellcolor[HTML]{ECF4FF} & \cellcolor[HTML]{ECF4FF}\textbf{73.58} & \cellcolor[HTML]{ECF4FF}\textbf{77.44} & \cellcolor[HTML]{ECF4FF} & \cellcolor[HTML]{ECF4FF}70.96 & \cellcolor[HTML]{ECF4FF}72.50 & \cellcolor[HTML]{ECF4FF} & \cellcolor[HTML]{ECF4FF}\textbf{72.63} & \cellcolor[HTML]{ECF4FF}\textbf{75.46} & \cellcolor[HTML]{ECF4FF} & \cellcolor[HTML]{ECF4FF}\textbf{77.00} & \cellcolor[HTML]{ECF4FF}\textbf{77.54} & \cellcolor[HTML]{ECF4FF} & \cellcolor[HTML]{ECF4FF}\textbf{73.54} & \cellcolor[HTML]{ECF4FF}\textbf{75.74} \\ \midrule
    
     & Self-Judge &  & 66.00 & 83.58 &  & 64.30 & 82.18 &  & 68.65 & 79.21 &  & 65.27 & 66.82 & & 66.06 & 77.95 \\
     & SelfCheckGPT &  & 72.15 & 86.12 &  & 50.75 & 69.70 &  & 67.73 & 77.31 &  & 75.57 & 76.11 & & 66.55 & 77.31 \\
     & Eigen-Score &  & 66.97 & 82.37 &  & 59.39 & 77.48 &  & 60.59 & 71.60 &  & 64.41 & 64.43 & & 62.84 & 73.97 \\
     & SAPLMA &  & 71.12 & 83.13 &  & 71.19 & 79.62 &  & 71.64 & \underline{78.62} &  & 77.17 & 77.22 & & 72.78 & 79.65 \\
     & {\hspace{15pt}\textit{+LaaB}} & \cellcolor[HTML]{ECF4FF} & \cellcolor[HTML]{ECF4FF}\underline{\textbf{73.20}} & \cellcolor[HTML]{ECF4FF}\underline{\textbf{86.22}} & \cellcolor[HTML]{ECF4FF} & \cellcolor[HTML]{ECF4FF}\underline{\textbf{71.40}} & \cellcolor[HTML]{ECF4FF}\underline{\textbf{81.75}} & \cellcolor[HTML]{ECF4FF} & \cellcolor[HTML]{ECF4FF}\underline{\textbf{71.71}} & \cellcolor[HTML]{ECF4FF}78.48 & \cellcolor[HTML]{ECF4FF} & \cellcolor[HTML]{ECF4FF}\underline{\textbf{79.15}} & \cellcolor[HTML]{ECF4FF}\underline{\textbf{79.44}} & \cellcolor[HTML]{ECF4FF} & \cellcolor[HTML]{ECF4FF}\underline{\textbf{73.87}} & \cellcolor[HTML]{ECF4FF}\underline{\textbf{81.47}} \\
     & Logits Lens &  & 65.67 & 76.55 &  & 62.42 & 70.63 &  & 54.50 & 60.03 &  & 72.33 & 72.52 & & 63.73 & 69.93 \\
     & {\hspace{15pt}\textit{+LaaB}} & \cellcolor[HTML]{ECF4FF} & \cellcolor[HTML]{ECF4FF}\textbf{70.23} & \cellcolor[HTML]{ECF4FF}\textbf{84.19} & \cellcolor[HTML]{ECF4FF} & \cellcolor[HTML]{ECF4FF}60.18 & \cellcolor[HTML]{ECF4FF}\textbf{72.33} & \cellcolor[HTML]{ECF4FF} & \cellcolor[HTML]{ECF4FF}\textbf{57.21} & \cellcolor[HTML]{ECF4FF}\textbf{71.74} & \cellcolor[HTML]{ECF4FF} & \cellcolor[HTML]{ECF4FF}72.07 & \cellcolor[HTML]{ECF4FF}72.36 & \cellcolor[HTML]{ECF4FF} & \cellcolor[HTML]{ECF4FF}\textbf{64.92} & \cellcolor[HTML]{ECF4FF}\textbf{75.16} \\
     & Lookback Lens &  & 69.97 & 79.53 &  & 68.95 & 75.07 &  & 64.30 & 68.67 &  & 78.31 & 78.54 & & 70.38 & 75.45 \\
    \multirow{-9}{*}{\makecell[l]{\textbf{Llama-3.1-}\\\textbf{70B-Instruct}}} & {\hspace{15pt}\textit{+LaaB}} & \cellcolor[HTML]{ECF4FF} & \cellcolor[HTML]{ECF4FF}\textbf{72.45} & \cellcolor[HTML]{ECF4FF}\textbf{85.11} & \cellcolor[HTML]{ECF4FF} & \cellcolor[HTML]{ECF4FF}\textbf{69.73} & \cellcolor[HTML]{ECF4FF}\textbf{79.62} & \cellcolor[HTML]{ECF4FF} & \cellcolor[HTML]{ECF4FF}\textbf{66.65} & \cellcolor[HTML]{ECF4FF}\textbf{76.28} & \cellcolor[HTML]{ECF4FF} & \cellcolor[HTML]{ECF4FF}\textbf{78.73} & \cellcolor[HTML]{ECF4FF}\textbf{78.91} & \cellcolor[HTML]{ECF4FF} & \cellcolor[HTML]{ECF4FF}\textbf{71.89} & \cellcolor[HTML]{ECF4FF}\textbf{79.98} \\ \midrule
    
     & Self-Judge &  & 72.19 & 78.73 &  & 65.14 & 77.54 &  & 71.29 & 73.99 &  & 69.07 & 69.28 & & 69.42 & 74.89 \\
     & SelfCheckGPT &  & 72.83 & 79.73 &  & 52.17 & 76.86 &  & 74.83 & 76.81 &  & 74.17 & 74.24 & & 68.50 & 76.91 \\
     & Eigen-Score &  & 67.52 & 73.51 &  & 56.98 & 75.35 &  & 61.33 & 62.63 &  & 69.19 & 69.24 & & 63.76 & 70.18 \\
     & SAPLMA &  & 76.73 & 80.09 &  & 69.38 & 78.38 &  & 76.81 & 77.55 &  & 76.65 & 76.76 & & 74.89 & 78.20 \\
     & {\hspace{15pt}\textit{+LaaB}} & \cellcolor[HTML]{ECF4FF} & \cellcolor[HTML]{ECF4FF}\underline{\textbf{79.42}} & \cellcolor[HTML]{ECF4FF}\underline{\textbf{83.85}} & \cellcolor[HTML]{ECF4FF} & \cellcolor[HTML]{ECF4FF}\textbf{69.60} & \cellcolor[HTML]{ECF4FF}\textbf{78.99} & \cellcolor[HTML]{ECF4FF} & \cellcolor[HTML]{ECF4FF}\underline{\textbf{79.35}} & \cellcolor[HTML]{ECF4FF}\underline{\textbf{80.65}} & \cellcolor[HTML]{ECF4FF} & \cellcolor[HTML]{ECF4FF}\textbf{77.17} & \cellcolor[HTML]{ECF4FF}\textbf{77.19} & \cellcolor[HTML]{ECF4FF} & \cellcolor[HTML]{ECF4FF}\underline{\textbf{76.39}} & \cellcolor[HTML]{ECF4FF}\underline{\textbf{80.17}} \\
     & Logits Lens &  & 68.15 & 72.74 &  & 53.19 & 63.17 &  & 65.02 & 66.03 &  & 72.56 & 72.62 & & 64.73 & 68.64 \\
     & {\hspace{15pt}\textit{+LaaB}} & \cellcolor[HTML]{ECF4FF} & \cellcolor[HTML]{ECF4FF}\textbf{70.90} & \cellcolor[HTML]{ECF4FF}\textbf{77.06} & \cellcolor[HTML]{ECF4FF} & \cellcolor[HTML]{ECF4FF}\textbf{54.62} & \cellcolor[HTML]{ECF4FF}\textbf{74.36} & \cellcolor[HTML]{ECF4FF} & \cellcolor[HTML]{ECF4FF}\textbf{67.21} & \cellcolor[HTML]{ECF4FF}\textbf{70.61} & \cellcolor[HTML]{ECF4FF} & \cellcolor[HTML]{ECF4FF}\textbf{73.80} & \cellcolor[HTML]{ECF4FF}\textbf{73.91} & \cellcolor[HTML]{ECF4FF} & \cellcolor[HTML]{ECF4FF}\textbf{66.63} & \cellcolor[HTML]{ECF4FF}\textbf{73.99} \\
     & Lookback Lens &  & 76.59 & 79.99 &  & 68.14 & 75.07 &  & 76.62 & 77.99 &  & \underline{78.46} & \underline{78.48} & & 74.95 & 77.88 \\
    \multirow{-9}{*}{\makecell[l]{\textbf{Qwen-2.5-}\\\textbf{32B-Instruct}}} & {\hspace{15pt}\textit{+LaaB}} & \cellcolor[HTML]{ECF4FF} & \cellcolor[HTML]{ECF4FF}\textbf{77.66} & \cellcolor[HTML]{ECF4FF}\textbf{82.36} & \cellcolor[HTML]{ECF4FF} & \cellcolor[HTML]{ECF4FF}\underline{\textbf{70.29}} & \cellcolor[HTML]{ECF4FF}\underline{\textbf{79.79}} & \cellcolor[HTML]{ECF4FF} & \cellcolor[HTML]{ECF4FF}\textbf{78.08} & \cellcolor[HTML]{ECF4FF}\textbf{79.32} & \cellcolor[HTML]{ECF4FF} & \cellcolor[HTML]{ECF4FF}78.16 & \cellcolor[HTML]{ECF4FF}78.21 & \cellcolor[HTML]{ECF4FF} & \cellcolor[HTML]{ECF4FF}\textbf{76.05} & \cellcolor[HTML]{ECF4FF}\textbf{79.92} \\ \midrule
    
     & Self-Judge &  & 56.84 & 73.55 &  & 42.28 & 61.80 &  & 46.86 & 64.81 &  & 43.10 & 49.54 & & 47.27 & 62.43 \\
     & SelfCheckGPT &  & 68.67 & 75.69 &  & 56.73 & 60.40 &  & 66.31 & 68.07 &  & 72.60 & 72.61 & & 66.08 & 69.19 \\
     & Eigen-Score &  & 67.78 & 72.06 &  & 55.84 & 59.33 &  & 59.14 & 62.67 &  & 65.11 & 65.35 & & 61.97 & 64.85 \\
     & SAPLMA &  & 76.72 & 79.02 &  & 70.89 & 71.36 &  & 72.96 & 74.36 &  & 74.87 & 75.00 & & 73.86 & 74.94 \\
     & {\hspace{15pt}\textit{+LaaB}} & \cellcolor[HTML]{ECF4FF} & \cellcolor[HTML]{ECF4FF}\underline{\textbf{77.69}} & \cellcolor[HTML]{ECF4FF}\underline{\textbf{80.40}} & \cellcolor[HTML]{ECF4FF} & \cellcolor[HTML]{ECF4FF}\textbf{70.99} & \cellcolor[HTML]{ECF4FF}71.26 & \cellcolor[HTML]{ECF4FF} & \cellcolor[HTML]{ECF4FF}\underline{\textbf{74.48}} & \cellcolor[HTML]{ECF4FF}\underline{\textbf{75.11}} & \cellcolor[HTML]{ECF4FF} & \cellcolor[HTML]{ECF4FF}\underline{\textbf{76.04}} & \cellcolor[HTML]{ECF4FF}\underline{\textbf{76.46}} & \cellcolor[HTML]{ECF4FF} & \cellcolor[HTML]{ECF4FF}\underline{\textbf{74.80}} & \cellcolor[HTML]{ECF4FF}\underline{\textbf{75.81}} \\
     & Logits Lens &  & 68.72 & 72.52 &  & 59.56 & 60.26 &  & 58.94 & 62.82 &  & 68.53 & 68.87 & & 63.94 & 66.12 \\
     & {\hspace{15pt}\textit{+LaaB}} & \cellcolor[HTML]{ECF4FF} & \cellcolor[HTML]{ECF4FF}67.73 & \cellcolor[HTML]{ECF4FF}\textbf{73.54} & \cellcolor[HTML]{ECF4FF} & \cellcolor[HTML]{ECF4FF}\textbf{60.98} & \cellcolor[HTML]{ECF4FF}\textbf{62.91} & \cellcolor[HTML]{ECF4FF} & \cellcolor[HTML]{ECF4FF}58.94 & \cellcolor[HTML]{ECF4FF}62.07 & \cellcolor[HTML]{ECF4FF} & \cellcolor[HTML]{ECF4FF}68.26 & \cellcolor[HTML]{ECF4FF}68.66 & \cellcolor[HTML]{ECF4FF} & \cellcolor[HTML]{ECF4FF}\textbf{63.98} & \cellcolor[HTML]{ECF4FF}\textbf{66.80} \\
     & Lookback Lens &  & 73.81 & 75.95 &  & 69.61 & 69.68 &  & 71.46 & 72.26 &  & 74.82 & 75.00 & & 72.43 & 73.22 \\
    \multirow{-9}{*}{\makecell[l]{\textbf{Mistral-7B-}\\\textbf{Instruct-v0.3}}} & {\hspace{15pt}\textit{+LaaB}} & \cellcolor[HTML]{ECF4FF} & \cellcolor[HTML]{ECF4FF}\textbf{74.46} & \cellcolor[HTML]{ECF4FF}\textbf{77.02} & \cellcolor[HTML]{ECF4FF} & \cellcolor[HTML]{ECF4FF}\underline{\textbf{71.24}} & \cellcolor[HTML]{ECF4FF}\underline{\textbf{72.10}} & \cellcolor[HTML]{ECF4FF} & \cellcolor[HTML]{ECF4FF}\textbf{71.67} & \cellcolor[HTML]{ECF4FF}\textbf{73.01} & \cellcolor[HTML]{ECF4FF} & \cellcolor[HTML]{ECF4FF}74.32 & \cellcolor[HTML]{ECF4FF}\textbf{75.11} & \cellcolor[HTML]{ECF4FF} & \cellcolor[HTML]{ECF4FF}\textbf{72.92} & \cellcolor[HTML]{ECF4FF}\textbf{74.31} \\ \bottomrule
    \end{tabular}\label{main results}
    \label{tab:main-results}
\end{table}

\noindent\textbf{Implementation Details.}
\label{sec:details}
For representation-based detectors, we configure the architectures of $D_r$ and $D_j$ as follows. For SAPLMA and Lookback Lens, both $D_r$ and $D_j$ employ a Multi-Layer Perceptron (MLP) classifier with hidden dimensions of $[256, 128, 64]$. For Logits Lens, we use a single Transformer layer with 4 attention heads to aggregate token-level features, followed by a MLP $D_r$ configured with $[64, 16]$ for classification; while $D_j$ utilizes an MLP with hidden dimensions of $[128, 64, 16]$. 
We select the optimal learning rate from $[1\text{e-}4, 5\text{e-}4, 1\text{e-}3, 5\text{e-}3]$ based on validation performance to conduct asynchronous training of $D_r$ and $D_j$ (the same lr selection criterion for baseline), followed by the joint fine-tuning with a learning rate of $1\text{e-}6$.
All classifiers are optimized using AdamW with a weight decay of 1e-5 and a dropout rate of 0.1. 
We adopt an early stopping strategy based on validation loss with a patience of 10 epochs.

For consistency-based detectors, each data instance is sampled 15 times using a decoding configuration of temperature=0.7, Top-p=0.9, and Top-k=10. Owing to the training-free attribute, we determine the optimal classification threshold via a grid search (step size = 0.1) on the validation set and report the results obtained on the test set.

\subsection{Main Results}

To evaluate the effectiveness of the \textbf{LaaB} framework, we conduct extensive experiments on four LLMs across four benchmarks. By comparing LaaB-enhanced detectors against various baselines, we aim to assess their capability in mining hallucination patterns by integrating information in two views guided by the logical constraints. The main results are summarized in \Tref{main results}. We observe that: 

\textbf{1) LaaB yields additional performance gains for the baselines using modeling intrinsic patterns in most cases.} 
The sustained improvement observed across most settings demonstrates the method's robust adaptability along two key dimensions. \textit{First}, LaaB exhibits exceptional generalization across diverse intrinsic patterns. As indicated by the \textbf{bold} values, it consistently augments detection capabilities when applied to hidden states (SAPLMA), logits (Logits Lens), and attention patterns (Lookback Lens). \textit{Second}, LaaB maintains its efficacy across varying model scales (ranging from 7B to 70B) and distinct architectures (including LLaMA, Qwen, and Mistral), underscoring its generalizability in capturing fundamental hallucination signatures across different LLM families.

\textbf{2) The hidden-state-based method SAPLMA shows higher detection performance than others and benefits more from LaaB.} 
Among base detectors, SAPLMA achieves higher detection performance than logits-based and attention-based methods in most cases. With the LaaB enhancement, SAPLMA shows more stable improvements and predominantly attains the best results across the majority of evaluation settings. 
This performance gap suggests that denser hidden-state representations possess a higher capacity for retaining informative signals, which are critical for effective hallucination detection. In contrast, logits-based detectors (Logits Lens) are limited by their inherent sparsity and discreteness, thus perform relatively worse among the three intrinsic pattern types.

\textbf{3) Detectors leveraging intrinsic patterns derived from LLMs' internal states outperform those leveraging the self-judgments or sampling estimates.} 
Detectors like SAPLMA, Logits Lens, and Lookback Lens show higher accuracy and macro F1 scores than Self-Judge and sampling-based baselines in most cases. This observation suggests that intrinsic model representations encode richer factuality-related signals than discrete token-level outputs, and based on that, the mutual learning from LaaB provides a more effective mechanism for exploiting such signals.

\begin{table}[htbp]
\centering\small
\captionsetup{justification=centering}
\caption{Performance comparison of LaaB and its variants on Llama-3.1-8B-Instruct}
\begin{tabular}{
      >{\centering\arraybackslash}m{1.6cm} 
      >{\raggedright\arraybackslash}m{2.6cm} 
      @{}c@{} 
      cc      
      >{\centering\arraybackslash}m{0.0cm} 
      cc      
      >{\centering\arraybackslash}m{0.0cm} 
      cc      
      >{\centering\arraybackslash}m{0.0cm} 
      cc      
    }
    \toprule
     &  & \textbf{} & \multicolumn{2}{c}{\textbf{TriviaQA}} & \textbf{} & \multicolumn{2}{c}{\textbf{MMLU}} & \textbf{} & \multicolumn{2}{c}{\textbf{NQ\_Open}} & \textbf{} & \multicolumn{2}{c}{\textbf{HaluEval}} \\ \cline{4-5} \cline{7-8} \cline{10-11} \cline{13-14} 
    \multirow{-2}{*}{\textbf{Method}} & \multirow{-2}{*}{\textbf{Variant}} &  & macF1 & Acc &  & macF1 & Acc &  & macF1 & Acc &  & macF1 & Acc \\ \midrule
    
    \multirow{3}{*}{\textbf{SAPLMA}} 
      & \cellcolor[HTML]{ECF4FF}LaaB & \cellcolor[HTML]{ECF4FF} & \cellcolor[HTML]{ECF4FF}78.74 & \cellcolor[HTML]{ECF4FF}82.09 & \cellcolor[HTML]{ECF4FF} & \cellcolor[HTML]{ECF4FF}71.77 & \cellcolor[HTML]{ECF4FF}73.25 & \cellcolor[HTML]{ECF4FF} & \cellcolor[HTML]{ECF4FF}73.10 & \cellcolor[HTML]{ECF4FF}77.13 & \cellcolor[HTML]{ECF4FF} & \cellcolor[HTML]{ECF4FF}77.20 & \cellcolor[HTML]{ECF4FF}77.60 \\ 
      & LaaB~($D_j$)  &  & 75.71 & 81.26 &  & 69.23 & 71.50 &  & 71.72 & 77.29 &  & 77.36 & 77.99 \\
      & LaaB~($D_r+D_j$) &  & 79.21 & 83.12 &  & 71.70 & 73.39 &  & 74.16 & 78.51 &  & 77.92 & 78.43 \\ \midrule

    \multirow{3}{*}{\makecell[c]{\textbf{Logits}\\\textbf{Lens}}} 
      & \cellcolor[HTML]{ECF4FF}LaaB & \cellcolor[HTML]{ECF4FF} & \cellcolor[HTML]{ECF4FF}75.11 & \cellcolor[HTML]{ECF4FF}78.88 & \cellcolor[HTML]{ECF4FF} & \cellcolor[HTML]{ECF4FF}66.72 & \cellcolor[HTML]{ECF4FF}70.65 & \cellcolor[HTML]{ECF4FF} & \cellcolor[HTML]{ECF4FF}65.72 & \cellcolor[HTML]{ECF4FF}68.90 & \cellcolor[HTML]{ECF4FF} & \cellcolor[HTML]{ECF4FF}72.29 & \cellcolor[HTML]{ECF4FF}73.04 \\ 
      & LaaB~($D_j$)  &  & 45.17 & 72.64 &  & 62.57 & 69.28 &  & 41.02 & 67.07 &  & 70.91 & 71.65 \\
      & LaaB~($D_r+D_j$) &  & 73.97 & 81.72 &  & 63.00 & 69.66 &  & 67.89 & 76.07 &  & 75.24 & 75.76 \\ \midrule

    \multirow{3}{*}{\makecell[c]{\textbf{Lookback}\\\textbf{Lens}}} 
      & \cellcolor[HTML]{ECF4FF}LaaB & \cellcolor[HTML]{ECF4FF} & \cellcolor[HTML]{ECF4FF}73.58 & \cellcolor[HTML]{ECF4FF}77.44 & \cellcolor[HTML]{ECF4FF} & \cellcolor[HTML]{ECF4FF}70.96 & \cellcolor[HTML]{ECF4FF}72.50 & \cellcolor[HTML]{ECF4FF} & \cellcolor[HTML]{ECF4FF}72.63 & \cellcolor[HTML]{ECF4FF}75.46 & \cellcolor[HTML]{ECF4FF} & \cellcolor[HTML]{ECF4FF}77.00 & \cellcolor[HTML]{ECF4FF}77.54 \\ 
      & LaaB~($D_j$)  &  & 69.42 & 78.63 &  & 61.89 & 69.19 &  & 67.87 & 75.61 &  & 71.99 & 73.10 \\
      & LaaB~($D_r+D_j$) &  & 75.26 & 80.74 &  & 63.69 & 69.90 &  & 72.24 & 77.44 &  & 76.61 & 77.43 \\ \bottomrule
\end{tabular}

\label{tab:ablation}
\end{table}

\subsection{Variant Analysis}

We evaluate two variants of LaaB to dissect the contributions of response and judgment hallucination detectors. \textbf{1)} \textit{LaaB~($D_j$)} uses only the judgment detector $D_j$ at inference to predict the self-judgment label and then derives the response-level decision according to the logical dependency shown in Table \ref{tab:label_logic}.
\textbf{2)} \textit{LaaB~($D_j+D_j$)} utilizes both the response and judgment detectors at inference and averages their scores under the logical constraints.

The results are shown in Table~\ref{tab:ablation}, LaaB mostly outperforms {LaaB~($D_j$)}, suggesting that the intrinsic patterns of responses remain the most informative cues for hallucination detection. $D_j$ primarily serves to complement $D_r$ by supplying additional signals from the self-judgment perspective. Moreover, LaaB~($D_j+D_j$) provides only marginal gains over LaaB despite nearly doubling LLM inference cost, indicating that most benefits of the judgment view are already distilled into the response detector $D_r$ during training. Overall, these ablations suggest that logic-constrained mutual learning successfully integrates both perspectives into the response detector, enabling efficient and effective hallucination detection using only $D_r$ at inference time.

\subsection{Cross-Dataset Generalization}

To assess the generalizability of our method under dataset shift, we adopt a leave-one-dataset-out evaluation protocol over four datasets. For each held-out benchmark, the hallucination detector is trained on the other three datasets and then evaluated on the held-out benchmark.

\begin{table}[htbp]
  \centering
  \caption{Leave-one-out Cross-Dataset Generalization Performance (macF1) on Llama-3.1-8B-Instruct. Both the baselines and LaaB are trained on the remaining three datasets and evaluated on the held-out benchmark.}
  \setlength{\tabcolsep}{7pt}
  \begin{tabular}{
    >{\raggedright\arraybackslash}c c c c c}
  \toprule
  \textbf{Method} & \textbf{TriviaQA} & \textbf{MMLU} & \textbf{NQ\_Open} & \textbf{HaluEval} \\
  \midrule
  SAPLMA          & 73.90 & 56.19 & 65.76 & 67.49 \\
  {\hspace{18pt}\textit{+LaaB}}
                  & \cellcolor[HTML]{ECF4FF}\textbf{\underline{78.18}}
                  & \cellcolor[HTML]{ECF4FF}\textbf{59.31}
                  & \cellcolor[HTML]{ECF4FF}\textbf{\underline{68.33}}
                  & \cellcolor[HTML]{ECF4FF}\textbf{\underline{70.25}}\\
  Logits Lens     & 73.05 & 56.65 & 57.25 & 67.78 \\
  {\hspace{18pt}\textit{+LaaB}}
                  & \cellcolor[HTML]{ECF4FF}\textbf{73.13}
                  & \cellcolor[HTML]{ECF4FF}55.05
                  & \cellcolor[HTML]{ECF4FF}\textbf{61.06}
                  & \cellcolor[HTML]{ECF4FF}\textbf{68.89}\\
  Lookback Lens   & 74.09 & 60.05 & 65.44 & 68.25  \\
  {\hspace{18pt}\textit{+LaaB}}
                  & \cellcolor[HTML]{ECF4FF}72.61
                  & \cellcolor[HTML]{ECF4FF}\textbf{\underline{63.61}}
                  & \cellcolor[HTML]{ECF4FF}\textbf{68.00}
                  & \cellcolor[HTML]{ECF4FF}\textbf{69.92}\\
  \bottomrule
  \end{tabular}
  \label{tab:crossdata}
\end{table}

As shown in Table~\ref{tab:crossdata}, equipping diverse hallucination detectors with LaaB consistently improves cross-dataset performance in most cases, suggesting enhanced robustness to distribution shift. One plausible explanation is that our logic-constrained mutual learning bridges two complementary signals—intrinsic prediction representations and self-judgment patterns. By encouraging logical agreement across two views, the training objective discourages reliance on spurious, dataset-specific cues that are unlikely to be supported by both signals, thereby pushing the detector toward evidence that transfers across datasets. Consequently, the learned decision boundary appears less sensitive to dataset-specific characteristics and transfers more reliably to unseen benchmarks.

\subsection{Further Analysis}
We perform further analysis to find out how LaaB brings detection improvement by breaking down the test set into subsets from different views. Specifically, we focus on the origin of the corrected instances and the length distribution. The following analysis is based on experiments with Llama-3.1-8B-Instruct, aggregated as the average across SAPLMA, Logits Lens, and Lookback Lens.

\begin{wrapfigure}{r}{0.48\textwidth}
  \centering
  \vspace{-1em}
  \includegraphics[width=0.45\textwidth]{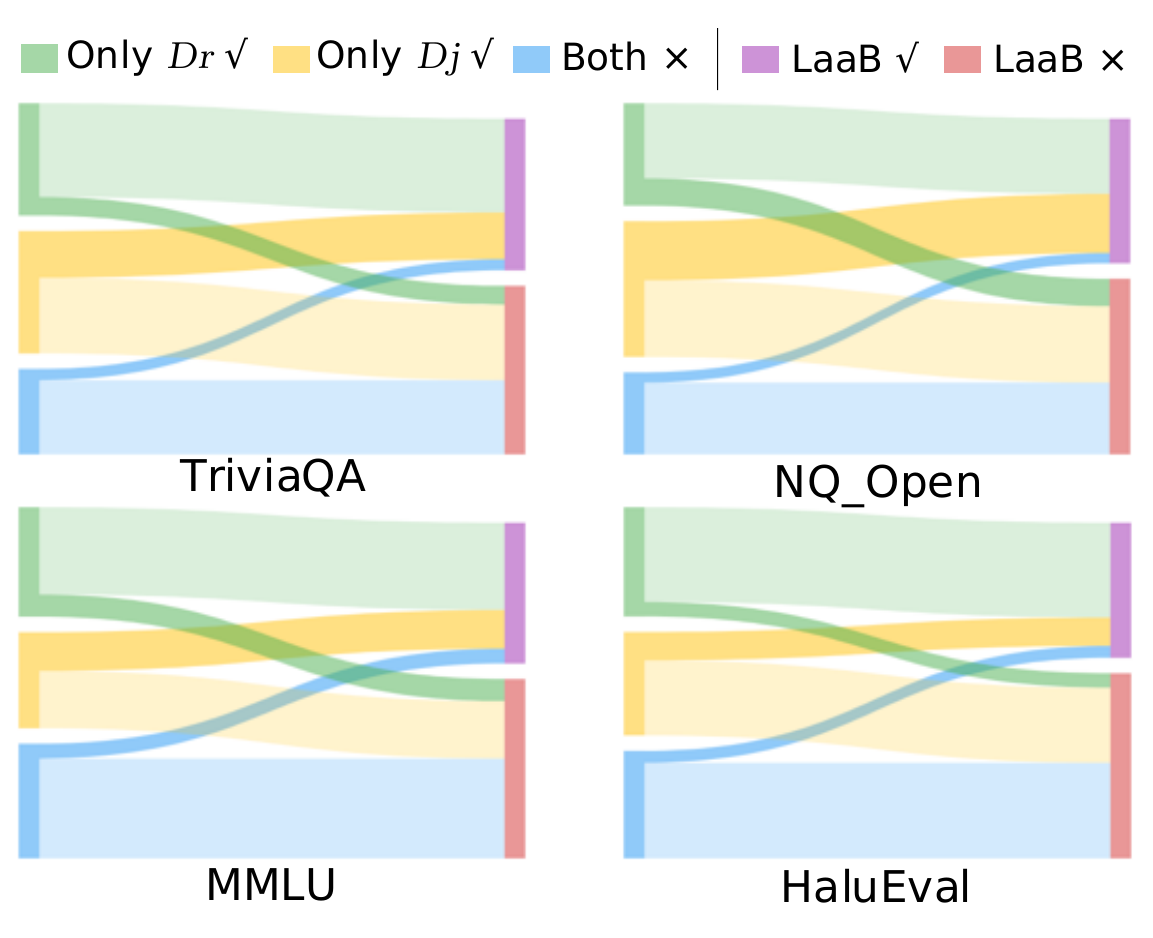}
  \caption{Prediction correctness transitions before and after applying LaaB.}
  \label{fig:sangji}
  \vspace{-1em}
\end{wrapfigure}

\paragraph{How do the predictions change before and after applying LaaB?} 
We categorized all testing data into four groups according to the independently trained detectors $D_r$ and $D_j$: Only $D_r$ $\checkmark$ (\textcolor{mygreen1}{\rule{1.5ex}{1.5ex}}), only $D_j$ $\checkmark$ (\textcolor{myyellow}{\rule{1.5ex}{1.5ex}}), both $\times$ (\textcolor{myblue1}{\rule{1.5ex}{1.5ex}}), and both $\checkmark$.
For the categories except ``both $\checkmark$'', we visualize the prediction correctness transition with the sankey diagram (\Fref{fig:sangji}), where flow widths are proportional to the absolute instance counts. We see that:

\textbf{1)} After adopting our proposed LaaB, a substantial portion of samples that $D_r$ originally predicted wrongly were corrected (the \colorbox{myyellow}{yellow} flow to \colorbox{mypurple}{purple} end), indicating that LaaB indeed transfers knowledge for the detection of hallucination from the self-judgment view to the response view.

\textbf{2)} Compared to the loss that $D_r$ turns into incorrect predictions (the \colorbox{mygreen1}{green} flow to \colorbox{myred}{red} end), LaaB preserves most correct predictions of $D_r$.

\begin{wrapfigure}{r}{0.48\textwidth}
  \centering
  \vspace{-1em}
  \includegraphics[width=0.45\textwidth]{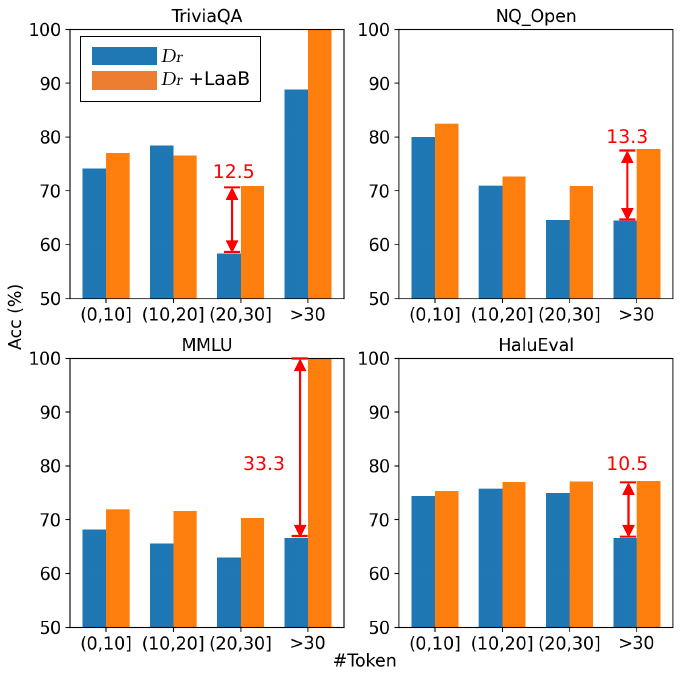}
  \caption{Performance on testing subsets with instances in different length intervals.}
  \label{fig:length-analysis}
  \vspace{-3em}
\end{wrapfigure}

\textbf{3)} Interestingly, we find a small but consistent flow from the ``both $\times$'' category to the correct class on all four datasets (the \colorbox{myblue1}{blue} flow to \colorbox{mypurple}{purple} end), which is beyond expectation. This might be because the logical constraints from LaaB introduced weak yet useful supervision signals that allow $D_r$ to refine the learned representations, even when both $D_r$ and $D_j$ predicted incorrectly before.

\paragraph{How effective is LaaB for instances with varying lengths?}
We calculate the accuracy of the subsets containing test instances in different length intervals. As presented in \Fref{fig:length-analysis}, LaaB brings accuracy improvement for most length intervals on all four datasets, demonstrating its general applicability. Among the subsets, we see a greater improvement at the long text intervals like $(20,30]$ and $> 30$. And the largest improvement occurs for the instances longer than 30 tokens at the MMLU dataset. This indicates that the hallucination detector can better handle more complex responses than before under the LaaB training. Inspired by \citet{prism}, a possible explanation is that self-judgment compresses response-level factuality into a single token (``Yes'' or ``No''), thereby mitigating representation sparsity and noise issues that arise with increasing sequence length.

\section{Conclusion}

We introduced LaaB (Logical Consistency-as-a-Bridge), a framework for LLM hallucination detection that bridges the gap between micro-level intrinsic neural patterns and macro-level symbolic self-judgments.
By introducing a ``meta-judgment'' process to map symbolic labels back into the feature space, LaaB provides a mutual learning framework between the hallucination detection modules for the response and the LLM self-judgment, guided by the inherent logical constraint. Extensive experiments on four public benchmarks, across four open-source LLMs, against eight baseline models show that LaaB can improve the hallucination detection performance for most base models, without introducing a significant increase in the inference cost. 

\noindent\textbf{Future Work.} We plan to further explore the following directions: 1) Evaluate LaaB and develop an improved version for hallucination detection in long articles; 2) Extend the core design to the multi-modality scenarios; and 3) Build a unified framework that integrates methods like LaaB with fact-checking pipelines.

\section*{Acknowledgment}

We adhere to the ACL Policy on Publication Ethics, including its Guidelines for Generative Assistance in Authorship~\citep{acl-policy}. We
used generative AI tools solely for language polishing of the manuscript and for assistance in drafting
certain portions of the code. All generated content
was reviewed and verified by the authors.

\section*{Limitations}
In this paper, we propose the integrated framework LaaB to combine the signals from LLMs' intrinsic patterns and self-judgments for hallucination detection. Despite its effectiveness, we identify the following limitations:

1) To obtain the self-judgment from the LLMs, we force the LLM to answer with ``Yes'' or ``No'', which is to avoid the situation where the LLM is originally intended to respond with ``I don't know.'' The compulsory constraint of the candidate answer space may bring some noise that negatively influences detectors' training.

2) Theoretically, our integration cannot correct the result of the samples that both intrinsic-pattern-based and self-judgment-based methods make incorrect predictions. The joint learning procedure may help some samples of this kind, but such an effect should be attributed to the joint optimization against ground-truth labels. Furthermore, our method does not guarantee the complete logical consistency for each sample in the resulting model because the applied constraint is soft.

3) The used intrinsic patterns and self-judgments are derived from the LLM that generates the given response, so our method is only applicable for the LLM service provider to monitor its service. The method cannot be used for third-party users who cannot access the internal information of LLMs. For these cases, fact-based misinformation detection~\cite{preffend} or black-box hallucination detection~\cite{bai2026inficheck} methods are more suitable.

\section*{Ethical Considerations}
\noindent\textbf{Risks.} Our work aims at detecting hallucinated LLM outputs and is suitable to be a monitoring component for LLM services, thus reducing the risk of users being misled. Given that an individual detector could hardly be perfect, real-world applications should consider multiple ways to detect hallucination more accurately.

\noindent\textbf{Data.} This work uses four publicly released datasets, including TriviaQA, MMLU, NQ\_Open, and HaluEval, under Apache-2.0 license, MIT license, Apache-2.0 license, and MIT license, respectively. 
We follow their intended use of academic research. During the research, we did not collect and use any unauthorized personal private data and did not recruit any human annotators.

\bibliography{custom}

@article{
wei2022h,
title={Emergent Abilities of Large Language Models},
author={Jason Wei and Yi Tay and Rishi Bommasani and Colin Raffel and Barret Zoph and Sebastian Borgeaud and Dani Yogatama and Maarten Bosma and Denny Zhou and Donald Metzler and Ed H. Chi and Tatsunori Hashimoto and Oriol Vinyals and Percy Liang and Jeff Dean and William Fedus},
journal={Transactions on Machine Learning Research},
issn={2835-8856},
year={2022},
url={https://openreview.net/forum?id=yzkSU5zdwD},
note={Survey Certification}
}

@article{zhang2025siren,
  title = {Siren’s Song in the AI Ocean: A Survey on Hallucination in Large Language Models},
  author = {Zhang, Yue and Li, Yafu and Cui, Leyang and Cai, Deng and Liu, Lemao and Fu, Tingchen and Huang, Xinting and Zhao, Enbo and Zhang, Yu and Chen, Yulong and Wang, Longyue and Luu, Anh Tuan and Bi, Wei and Shi, Freda and Shi, Shuming},
    journal = {Computational Linguistics},
    volume = {51},
    number = {4},
    pages = {1373--1418},
    year = {2025},
    month = {12},
  publisher={MIT Press},
  doi={10.1162/COLI.a.16}
}

@article{ji2023survey,
author = {Ji, Ziwei and Lee, Nayeon and Frieske, Rita and Yu, Tiezheng and Su, Dan and Xu, Yan and Ishii, Etsuko and Bang, Ye Jin and Madotto, Andrea and Fung, Pascale},
title = {Survey of Hallucination in Natural Language Generation},
year = {2023},
issue_date = {December 2023},
publisher = {Association for Computing Machinery},
volume = {55},
number = {12},
issn = {0360-0300},
doi = {10.1145/3571730},
journal = {ACM Comput. Surv.},
month = mar,
articleno = {248},
numpages = {38},
}

@article{huang2025survey,
author = {Huang, Lei and Yu, Weijiang and Ma, Weitao and Zhong, Weihong and Feng, Zhangyin and Wang, Haotian and Chen, Qianglong and Peng, Weihua and Feng, Xiaocheng and Qin, Bing and Liu, Ting},
title = {A Survey on Hallucination in Large Language Models: Principles, Taxonomy, Challenges, and Open Questions},
year = {2025},
issue_date = {March 2025},
publisher = {Association for Computing Machinery},
volume = {43},
number = {2},
issn = {1046-8188},
url = {https://doi.org/10.1145/3703155},
doi = {10.1145/3703155},
journal = {ACM Trans. Inf. Syst.},
month = jan,
articleno = {42},
numpages = {55},
}

@misc{mohsin2025fundamental,
      title={On the Fundamental Limits of LLMs at Scale}, 
      author={Muhammad Ahmed Mohsin and Muhammad Umer and Ahsan Bilal and Zeeshan Memon and Muhammad Ibtsaam Qadir and Sagnik Bhattacharya and Hassan Rizwan and Abhiram R. Gorle and Maahe Zehra Kazmi and Nukhba Amir and Ali Subhan and Muhammad Usman Rafique and Zihao He and Pulkit Mehta and Muhammad Ali Jamshed and John M. Cioffi},
      year={2026},
      eprint={2511.12869},
      archivePrefix={arXiv},
      primaryClass={cs.LG},
      url={https://arxiv.org/abs/2511.12869}, 
}

@misc{xu2024hallucination,
      title={Hallucination is Inevitable: An Innate Limitation of Large Language Models}, 
      author={Ziwei Xu and Sanjay Jain and Mohan Kankanhalli},
      year={2025},
      eprint={2401.11817},
      archivePrefix={arXiv},
      primaryClass={cs.CL},
      url={https://arxiv.org/abs/2401.11817}, 
}

@article{panickssery2024llm,
  title={Llm evaluators recognize and favor their own generations},
  author={Panickssery, Arjun and Bowman, Samuel and Feng, Shi},
  journal={Advances in Neural Information Processing Systems},
  volume={37},
  pages={68772--68802},
  year={2024},
  doi={10.52202/079017-2197}
}

@inproceedings{
wataoka2024self,
title={Self-Preference Bias in {LLM}-as-a-Judge},
author={Koki Wataoka and Tsubasa Takahashi and Ryokan Ri},
booktitle={Neurips Safe Generative AI Workshop 2024},
year={2024},
url={https://openreview.net/forum?id=tLZZZIgPJX}
}

@inproceedings{tan2025too,
    title = "Too Consistent to Detect: A Study of Self-Consistent Errors in {LLM}s",
    author = "Tan, Hexiang  and
      Sun, Fei  and
      Liu, Sha  and
      Su, Du  and
      Cao, Qi  and
      Chen, Xin  and
      Wang, Jingang  and
      Cai, Xunliang  and
      Wang, Yuanzhuo  and
      Shen, Huawei  and
      Cheng, Xueqi",
    booktitle = "Proceedings of the 2025 Conference on Empirical Methods in Natural Language Processing",
    month = nov,
    year = "2025",
    publisher = "Association for Computational Linguistics",
    url = "https://aclanthology.org/2025.emnlp-main.238/",
    doi = "10.18653/v1/2025.emnlp-main.238",
    pages = "4755--4765",
    ISBN = "979-8-89176-332-6",
}

@inproceedings{zhou2024relying,
    title = "Relying on the Unreliable: The Impact of Language Models' Reluctance to Express Uncertainty",
    author = "Zhou, Kaitlyn  and
      Hwang, Jena D.  and
      Ren, Xiang  and
      Sap, Maarten",
    booktitle = "Proceedings of the 62nd Annual Meeting of the Association for Computational Linguistics (Volume 1: Long Papers)",
    month = aug,
    year = "2024",
    publisher = "Association for Computational Linguistics",
    url = "https://aclanthology.org/2024.acl-long.198/",
    doi = "10.18653/v1/2024.acl-long.198",
    pages = "3623--3643",
}

@inproceedings{
wen2024mitigating,
title={Mitigating overconfidence in large language models: A behavioral lens on confidence estimation and calibration},
author={Bingbing Wen and Chenjun Xu and Bin HAN and Robert Wolfe and Lucy Lu Wang and Bill Howe},
booktitle={NeurIPS 2024 Workshop on Behavioral Machine Learning},
year={2024},
url={https://openreview.net/forum?id=y9UdO5cmHs}
}

@inproceedings{simhi2025trust,
    title = "Trust Me, {I}{'}m Wrong: {LLM}s Hallucinate with Certainty Despite Knowing the Answer",
    author = "Simhi, Adi  and
      Itzhak, Itay  and
      Barez, Fazl  and
      Stanovsky, Gabriel  and
      Belinkov, Yonatan",
    booktitle = "Findings of the Association for Computational Linguistics: EMNLP 2025",
    month = nov,
    year = "2025",
    publisher = "Association for Computational Linguistics",
    url = "https://aclanthology.org/2025.findings-emnlp.792/",
    doi = "10.18653/v1/2025.findings-emnlp.792",
    pages = "14665--14688",
    ISBN = "979-8-89176-335-7"
}

@inproceedings{
park2025steer,
title={Steer {LLM} Latents for Hallucination Detection},
author={Seongheon Park and Xuefeng Du and Min-Hsuan Yeh and Haobo Wang and Yixuan Li},
booktitle={Forty-second International Conference on Machine Learning},
year={2025},
url={https://openreview.net/forum?id=UMqNQEPNT3}
}

@inproceedings{manakul2023selfcheckgpt,
    title = "{S}elf{C}heck{GPT}: Zero-Resource Black-Box Hallucination Detection for Generative Large Language Models",
    author = "Manakul, Potsawee  and
      Liusie, Adian  and
      Gales, Mark",
    booktitle = "Proceedings of the 2023 Conference on Empirical Methods in Natural Language Processing",
    month = dec,
    year = "2023",
    publisher = "Association for Computational Linguistics",
    url = "https://aclanthology.org/2023.emnlp-main.557/",
    doi = "10.18653/v1/2023.emnlp-main.557",
    pages = "9004--9017",
}

@inproceedings{guo2017calibration,
  title={On calibration of modern neural networks},
  author={Guo, Chuan and Pleiss, Geoff and Sun, Yu and Weinberger, Kilian Q},
  booktitle={International conference on machine learning},
  pages={1321--1330},
  year={2017},
  organization={PMLR},
  url={https://proceedings.mlr.press/v70/guo17a/guo17a.pdf}
}

@misc{ma2025estimating,
      title={Estimating LLM Uncertainty with Evidence}, 
      author={Huan Ma and Jingdong Chen and Joey Tianyi Zhou and Guangyu Wang and Changqing Zhang},
      year={2025},
      eprint={2502.00290},
      archivePrefix={arXiv},
      primaryClass={cs.CL},
      url={https://arxiv.org/abs/2502.00290}, 
}

@inproceedings{azaria2023internal,
    title = "The Internal State of an {LLM} Knows When It{'}s Lying",
    author = "Azaria, Amos  and
      Mitchell, Tom",
    booktitle = "Findings of the Association for Computational Linguistics: EMNLP 2023",
    month = dec,
    year = "2023",
    publisher = "Association for Computational Linguistics",
    url = "https://aclanthology.org/2023.findings-emnlp.68/",
    doi = "10.18653/v1/2023.findings-emnlp.68",
    pages = "967--976",
}

@inproceedings{chuang2024lookback,
    title = "Lookback Lens: Detecting and Mitigating Contextual Hallucinations in Large Language Models Using Only Attention Maps",
    author = "Chuang, Yung-Sung  and
      Qiu, Linlu  and
      Hsieh, Cheng-Yu  and
      Krishna, Ranjay  and
      Kim, Yoon  and
      Glass, James R.",
    booktitle = "Proceedings of the 2024 Conference on Empirical Methods in Natural Language Processing",
    month = nov,
    year = "2024",
    publisher = "Association for Computational Linguistics",
    url = "https://aclanthology.org/2024.emnlp-main.84/",
    doi = "10.18653/v1/2024.emnlp-main.84",
    pages = "1419--1436",
}

@inproceedings{ji2023towards,
    title = "Towards Mitigating {LLM} Hallucination via Self Reflection",
    author = "Ji, Ziwei  and
      Yu, Tiezheng  and
      Xu, Yan  and
      Lee, Nayeon  and
      Ishii, Etsuko  and
      Fung, Pascale",
    booktitle = "Findings of the Association for Computational Linguistics: EMNLP 2023",
    month = dec,
    year = "2023",
    publisher = "Association for Computational Linguistics",
    url = "https://aclanthology.org/2023.findings-emnlp.123/",
    doi = "10.18653/v1/2023.findings-emnlp.123",
    pages = "1827--1843",
}

@inproceedings{li2025hallucination,
    title = "Hallucination Detection in Structured Query Generation via {LLM} Self-Debating",
    author = "Li, Miaoran  and
      Chen, Jiangning  and
      Xu, Minghua  and
      Wang, Xiaolong",
    booktitle = "Findings of the Association for Computational Linguistics: EMNLP 2025",
    month = nov,
    year = "2025",
    publisher = "Association for Computational Linguistics",
    url = "https://aclanthology.org/2025.findings-emnlp.873/",
    doi = "10.18653/v1/2025.findings-emnlp.873",
    pages = "16102--16113",
    ISBN = "979-8-89176-335-7",
}

@inproceedings{zhang2025understanding,
    title = "Understanding the Dark Side of {LLM}s' Intrinsic Self-Correction",
    author = "Zhang, Qingjie  and
      Wang, Di  and
      Qian, Haoting  and
      Li, Yiming  and
      Zhang, Tianwei  and
      Huang, Minlie  and
      Xu, Ke  and
      Li, Hewu  and
      Yan, Liu  and
      Qiu, Han",
    booktitle = "Proceedings of the 63rd Annual Meeting of the Association for Computational Linguistics (Volume 1: Long Papers)",
    month = jul,
    year = "2025",
    publisher = "Association for Computational Linguistics",
    url = "https://aclanthology.org/2025.acl-long.1314/",
    doi = "10.18653/v1/2025.acl-long.1314",
    pages = "27066--27101",
    ISBN = "979-8-89176-251-0",
}

@misc{su2025between,
      title={Between Underthinking and Overthinking: An Empirical Study of Reasoning Length and correctness in LLMs}, 
      author={Jinyan Su and Jennifer Healey and Preslav Nakov and Claire Cardie},
      year={2025},
      eprint={2505.00127},
      archivePrefix={arXiv},
      primaryClass={cs.CL},
      url={https://arxiv.org/abs/2505.00127}, 
}

@article{farquhar2024detecting,
  title={Detecting hallucinations in large language models using semantic entropy},
  author={Farquhar, Sebastian and Kossen, Jannik and Kuhn, Lorenz and Gal, Yarin},
  journal={Nature},
  volume={630},
  pages={625--630},
  year={2024},
  publisher={Nature Publishing Group},
  doi={10.1038/s41586-024-07421-0}
}

@inproceedings{yin2023large,
    title = "Do Large Language Models Know What They Don{'}t Know?",
    author = "Yin, Zhangyue  and
      Sun, Qiushi  and
      Guo, Qipeng  and
      Wu, Jiawen  and
      Qiu, Xipeng  and
      Huang, Xuanjing",
    booktitle = "Findings of the Association for Computational Linguistics: ACL 2023",
    month = jul,
    year = "2023",
    publisher = "Association for Computational Linguistics",
    doi = "10.18653/v1/2023.findings-acl.551",
    pages = "8653--8665",
}

@inproceedings{
xiongcan,
title={Can {LLM}s Express Their Uncertainty? An Empirical Evaluation of Confidence Elicitation in {LLM}s},
author={Miao Xiong and Zhiyuan Hu and Xinyang Lu and YIFEI LI and Jie Fu and Junxian He and Bryan Hooi},
booktitle={The Twelfth International Conference on Learning Representations},
year={2024},
url={https://openreview.net/forum?id=gjeQKFxFpZ}
}

@inproceedings{liu2025long,
    title = "Long-form Hallucination Detection with Self-elicitation",
    author = "Liu, Zihang  and
      Guo, Jiawei  and
      Zhang, Hao  and
      Chen, Hongyang  and
      Bu, Jiajun  and
      Wang, Haishuai",
    booktitle = "Findings of the Association for Computational Linguistics: ACL 2025",
    month = jul,
    year = "2025",
    publisher = "Association for Computational Linguistics",
    url = "https://aclanthology.org/2025.findings-acl.211/",
    doi = "10.18653/v1/2025.findings-acl.211",
    pages = "4082--4100",
    ISBN = "979-8-89176-256-5",
}

@inproceedings{zhang2018deep,
  title={Deep mutual learning},
  author={Zhang, Ying and Xiang, Tao and Hospedales, Timothy M and Lu, Huchuan},
  booktitle={Proceedings of the IEEE/CVF Conference on Computer Vision and Pattern Recognition}, 
  pages={4320--4328},
  year={2018},
  doi={10.1109/CVPR.2018.00454}
}

@inproceedings{guo2020online,
  title={Online knowledge distillation via collaborative learning},
  author={Guo, Qiushan and Wang, Xinjiang and Wu, Yichao and Yu, Zhipeng and Liang, Ding and Hu, Xiaolin and Luo, Ping},
  booktitle={Proceedings of the IEEE/CVF conference on computer vision and pattern recognition},
  pages={11017-11026},
  year={2020},
  DOI={10.1109/CVPR42600.2020.01103}
}

@article{tan2022online,
  title={Online knowledge distillation with elastic peer},
  author={Tan, Chao and Liu, Jie},
  journal={Information Sciences},
  volume={583},
  pages={1--13},
  year={2022},
  publisher={Elsevier},
  DOI={10.1016/j.ins.2021.10.043}
}

@article{yang2023online,
  title={Online knowledge distillation via mutual contrastive learning for visual recognition},
  author={Yang, Chuanguang and An, Zhulin and Zhou, Helong and Zhuang, Fuzhen and Xu, Yongjun and Zhang, Qian},
  journal={IEEE Transactions on Pattern Analysis and Machine Intelligence},
  volume={45},
  number={8},
  pages={10212--10227},
  year={2023},
  publisher={IEEE},
  DOI={10.1109/TPAMI.2023.3257878}
}

@inproceedings{liu2024universal,
    title = "On the Universal Truthfulness Hyperplane Inside {LLM}s",
    author = "Liu, Junteng  and
      Chen, Shiqi  and
      Cheng, Yu  and
      He, Junxian",
    booktitle = "Proceedings of the 2024 Conference on Empirical Methods in Natural Language Processing",
    month = nov,
    year = "2024",
    publisher = "Association for Computational Linguistics",
    url = "https://aclanthology.org/2024.emnlp-main.1012/",
    doi = "10.18653/v1/2024.emnlp-main.1012",
    pages = "18199--18224",
}

@inproceedings{su2024unsupervised,
    title = "Unsupervised Real-Time Hallucination Detection based on the Internal States of Large Language Models",
    author = "Su, Weihang  and
      Wang, Changyue  and
      Ai, Qingyao  and
      Hu, Yiran  and
      Wu, Zhijing  and
      Zhou, Yujia  and
      Liu, Yiqun",
    booktitle = "Findings of the Association for Computational Linguistics: ACL 2024",
    month = aug,
    year = "2024",
    publisher = "Association for Computational Linguistics",
    url = "https://aclanthology.org/2024.findings-acl.854/",
    doi = "10.18653/v1/2024.findings-acl.854",
    pages = "14379--14391",
}

@article{du2024haloscope,
  title={Haloscope: Harnessing unlabeled LLM generations for hallucination detection},
  author={Du, Xuefeng and Xiao, Chaowei and Li, Yixuan},
  journal={Advances in Neural Information Processing Systems},
  volume={37},
  pages={102948--102972},
  year={2024},
  publisher = {Curran Associates, Inc.},
  doi={10.52202/079017-3270}
}

@inproceedings{
wanglatent,
title={Latent Space Chain-of-Embedding Enables Output-free {LLM} Self-Evaluation},
author={Yiming Wang and Pei Zhang and Baosong Yang and Derek F. Wong and Rui Wang},
booktitle={The Thirteenth International Conference on Learning Representations},
year={2025},
url={https://openreview.net/forum?id=jxo70B9fQo}
}

@inproceedings{vashurin2025uncertainty,
    title = "{UNCERTAINTY}-{LINE}: Length-Invariant Estimation of Uncertainty for Large Language Models",
    author = "Vashurin, Roman  and
      Goloburda, Maiya  and
      Nakov, Preslav  and
      Panov, Maxim",
    booktitle = "Proceedings of the 2025 Conference on Empirical Methods in Natural Language Processing",
    month = nov,
    year = "2025",
    publisher = "Association for Computational Linguistics",
    url = "https://aclanthology.org/2025.emnlp-main.400/",
    doi = "10.18653/v1/2025.emnlp-main.400",
    pages = "7881--7908",
    ISBN = "979-8-89176-332-6",
}

@inproceedings{vazhentsev2025unconditional,
    title = "Unconditional Truthfulness: Learning Unconditional Uncertainty of Large Language Models",
    author = "Vazhentsev, Artem  and
      Fadeeva, Ekaterina  and
      Xing, Rui  and
      Kuzmin, Gleb  and
      Lazichny, Ivan  and
      Panchenko, Alexander  and
      Nakov, Preslav  and
      Baldwin, Timothy  and
      Panov, Maxim  and
      Shelmanov, Artem",
    booktitle = "Proceedings of the 2025 Conference on Empirical Methods in Natural Language Processing",
    month = nov,
    year = "2025",
    publisher = "Association for Computational Linguistics",
    url = "https://aclanthology.org/2025.emnlp-main.1807/",
    doi = "10.18653/v1/2025.emnlp-main.1807",
    pages = "35673--35694",
    ISBN = "979-8-89176-332-6",
}

@inproceedings{jiang2024large,
    title = "On Large Language Models' Hallucination with Regard to Known Facts",
    author = "Jiang, Che  and
      Qi, Biqing  and
      Hong, Xiangyu  and
      Fu, Dayuan  and
      Cheng, Yang  and
      Meng, Fandong  and
      Yu, Mo  and
      Zhou, Bowen  and
      Zhou, Jie",
    booktitle = "Proceedings of the 2024 Conference of the North American Chapter of the Association for Computational Linguistics: Human Language Technologies (Volume 1: Long Papers)",
    month = jun,
    year = "2024",
    publisher = "Association for Computational Linguistics",
    url = "https://aclanthology.org/2024.naacl-long.60/",
    doi = "10.18653/v1/2024.naacl-long.60",
    pages = "1041--1053",
}

@inproceedings{liu2025attention,
    title = "Attention-guided Self-reflection for Zero-shot Hallucination Detection in Large Language Models",
    author = "Liu, Qiang  and
      Chen, Xinlong  and
      Ding, Yue  and
      Song, Bowen  and
      Wang, Weiqiang  and
      Wu, Shu  and
      Wang, Liang",
    booktitle = "Proceedings of the 2025 Conference on Empirical Methods in Natural Language Processing",
    month = nov,
    year = "2025",
    publisher = "Association for Computational Linguistics",
    url = "https://aclanthology.org/2025.emnlp-main.1063/",
    doi = "10.18653/v1/2025.emnlp-main.1063",
    pages = "21005--21021",
    ISBN = "979-8-89176-332-6",
}

@inproceedings{binkowski2025hallucination,
    title = "Hallucination Detection in {LLM}s Using Spectral Features of Attention Maps",
    author = "Binkowski, Jakub  and
      Janiak, Denis  and
      Sawczyn, Albert  and
      Gabrys, Bogdan  and
      Kajdanowicz, Tomasz Jan",
    booktitle = "Proceedings of the 2025 Conference on Empirical Methods in Natural Language Processing",
    month = nov,
    year = "2025",
    publisher = "Association for Computational Linguistics",
    url = "https://aclanthology.org/2025.emnlp-main.1239/",
    doi = "10.18653/v1/2025.emnlp-main.1239",
    pages = "24354--24385",
    ISBN = "979-8-89176-332-6",
}

@inproceedings{zhang2025icr,
    title = "{ICR} Probe: Tracking Hidden State Dynamics for Reliable Hallucination Detection in {LLM}s",
    author = "Zhang, Zhenliang  and
      Hu, Xinyu  and
      Zhang, Huixuan  and
      Zhang, Junzhe  and
      Wan, Xiaojun",
    booktitle = "Proceedings of the 63rd Annual Meeting of the Association for Computational Linguistics (Volume 1: Long Papers)",
    month = jul,
    year = "2025",
    publisher = "Association for Computational Linguistics",
    url = "https://aclanthology.org/2025.acl-long.880/",
    doi = "10.18653/v1/2025.acl-long.880",
    pages = "17986--18002",
    ISBN = "979-8-89176-251-0",
}

@inproceedings{hu-etal-2024-knowledge,
    title = "Knowledge-Centric Hallucination Detection",
    author = "Hu, Xiangkun  and
      Ru, Dongyu  and
      Qiu, Lin  and
      Guo, Qipeng  and
      Zhang, Tianhang  and
      Xu, Yang  and
      Luo, Yun  and
      Liu, Pengfei  and
      Zhang, Yue  and
      Zhang, Zheng",
    booktitle = "Proceedings of the 2024 Conference on Empirical Methods in Natural Language Processing",
    month = nov,
    year = "2024",
    publisher = "Association for Computational Linguistics",
    doi = "10.18653/v1/2024.emnlp-main.395",
    pages = "6953--6975",
}

@misc{kadavath2022language,
      title={Language Models (Mostly) Know What They Know}, 
      author={Saurav Kadavath and Tom Conerly and Amanda Askell and Tom Henighan and Dawn Drain and Ethan Perez and Nicholas Schiefer and Zac Hatfield-Dodds and Nova DasSarma and Eli Tran-Johnson and Scott Johnston and Sheer El-Showk and Andy Jones and Nelson Elhage and Tristan Hume and Anna Chen and Yuntao Bai and Sam Bowman and Stanislav Fort and Deep Ganguli and Danny Hernandez and Josh Jacobson and Jackson Kernion and Shauna Kravec and Liane Lovitt and Kamal Ndousse and Catherine Olsson and Sam Ringer and Dario Amodei and Tom Brown and Jack Clark and Nicholas Joseph and Ben Mann and Sam McCandlish and Chris Olah and Jared Kaplan},
      year={2022},
      eprint={2207.05221},
      archivePrefix={arXiv},
      primaryClass={cs.CL},
      url={https://arxiv.org/abs/2207.05221}, 
}

@inproceedings{shi2024ten,
  title={Ten words only still help: improving black-box AI-generated text detection via proxy-guided efficient re-sampling},
  author={Shi, Yuhui and Sheng, Qiang and Cao, Juan and Mi, Hao and Hu, Beizhe and Wang, Danding},
  booktitle={Proceedings of the Thirty-Third International Joint Conference on Artificial Intelligence},
  pages={494--502},
  year={2024},
  DOI={10.24963/ijcai.2024/55}
}

@inproceedings{wang2023seqxgpt,
    title = "{S}eq{XGPT}: Sentence-Level {AI}-Generated Text Detection",
    author = "Wang, Pengyu  and
      Li, Linyang  and
      Ren, Ke  and
      Jiang, Botian  and
      Zhang, Dong  and
      Qiu, Xipeng",
    booktitle = "Proceedings of the 2023 Conference on Empirical Methods in Natural Language Processing",
    month = dec,
    year = "2023",
    publisher = "Association for Computational Linguistics",
    url = "https://aclanthology.org/2023.emnlp-main.73/",
    pages = "1144--1156",
}

@misc{acl-policy,
    author={Aoife Cahill and Leon Derczynski and Kokil Jaidka},
    title={{ACL Policy on Publication Ethics}},
    year={2025},
    url={https://www.aclweb.org/adminwiki/index.php/ACL_Policy_on_Publication_Ethics#Guidelines_for_Generative_Assistance_in_Authorship},
    note={Accessed: 2026-01-02}
}

@misc{logit-lens,
    author={nostalgebraist},
    title={{interpreting GPT: the logit lens}},
    year={2020},
    url={https://www.lesswrong.com/posts/AcKRB8wDpdaN6v6ru/interpreting-gpt-the-logit-lens},
    note={Accessed: 2026-01-02}
}

@misc{llama3,
      title={The Llama 3 Herd of Models}, 
      author={Aaron Grattafiori and Abhimanyu Dubey and Abhinav Jauhri and Abhinav Pandey and Abhishek Kadian and Ahmad Al-Dahle and Aiesha Letman and Akhil Mathur and Alan Schelten and Alex Vaughan and Amy Yang and Angela Fan and Anirudh Goyal and Anthony Hartshorn and Aobo Yang and Archi Mitra and Archie Sravankumar and Artem Korenev and Arthur Hinsvark and Arun Rao and Aston Zhang and Aurelien Rodriguez and Austen Gregerson and Ava Spataru and Baptiste Roziere and Bethany Biron and Binh Tang and Bobbie Chern and Charlotte Caucheteux and Chaya Nayak and Chloe Bi and Chris Marra and Chris McConnell and Christian Keller and Christophe Touret and Chunyang Wu and Corinne Wong and Cristian Canton Ferrer and Cyrus Nikolaidis and Damien Allonsius and Daniel Song and Danielle Pintz and Danny Livshits and Danny Wyatt and David Esiobu and Dhruv Choudhary and Dhruv Mahajan and Diego Garcia-Olano and Diego Perino and Dieuwke Hupkes and Egor Lakomkin and Ehab AlBadawy and Elina Lobanova and Emily Dinan and Eric Michael Smith and Filip Radenovic and Francisco Guzmán and Frank Zhang and Gabriel Synnaeve and Gabrielle Lee and Georgia Lewis Anderson and Govind Thattai and Graeme Nail and Gregoire Mialon and Guan Pang and Guillem Cucurell and Hailey Nguyen and Hannah Korevaar and Hu Xu and Hugo Touvron and Iliyan Zarov and Imanol Arrieta Ibarra and Isabel Kloumann and Ishan Misra and Ivan Evtimov and Jack Zhang and Jade Copet and Jaewon Lee and Jan Geffert and Jana Vranes and Jason Park and Jay Mahadeokar and Jeet Shah and Jelmer van der Linde and Jennifer Billock and Jenny Hong and Jenya Lee and Jeremy Fu and Jianfeng Chi and Jianyu Huang and Jiawen Liu and Jie Wang and Jiecao Yu and Joanna Bitton and Joe Spisak and Jongsoo Park and Joseph Rocca and Joshua Johnstun and Joshua Saxe and Junteng Jia and Kalyan Vasuden Alwala and Karthik Prasad and Kartikeya Upasani and Kate Plawiak and Ke Li and Kenneth Heafield and Kevin Stone and Khalid El-Arini and Krithika Iyer and Kshitiz Malik and Kuenley Chiu and Kunal Bhalla and Kushal Lakhotia and Lauren Rantala-Yeary and Laurens van der Maaten and Lawrence Chen and Liang Tan and Liz Jenkins and Louis Martin and Lovish Madaan and Lubo Malo and Lukas Blecher and Lukas Landzaat and Luke de Oliveira and Madeline Muzzi and Mahesh Pasupuleti and Mannat Singh and Manohar Paluri and Marcin Kardas and Maria Tsimpoukelli and Mathew Oldham and Mathieu Rita and Maya Pavlova and Melanie Kambadur and Mike Lewis and Min Si and Mitesh Kumar Singh and Mona Hassan and Naman Goyal and Narjes Torabi and Nikolay Bashlykov and Nikolay Bogoychev and Niladri Chatterji and Ning Zhang and Olivier Duchenne and Onur Çelebi and Patrick Alrassy and Pengchuan Zhang and Pengwei Li and Petar Vasic and Peter Weng and Prajjwal Bhargava and Pratik Dubal and Praveen Krishnan and Punit Singh Koura and Puxin Xu and Qing He and Qingxiao Dong and Ragavan Srinivasan and Raj Ganapathy and Ramon Calderer and Ricardo Silveira Cabral and Robert Stojnic and Roberta Raileanu and Rohan Maheswari and Rohit Girdhar and Rohit Patel and Romain Sauvestre and Ronnie Polidoro and Roshan Sumbaly and Ross Taylor and Ruan Silva and Rui Hou and Rui Wang and Saghar Hosseini and Sahana Chennabasappa and Sanjay Singh and Sean Bell and Seohyun Sonia Kim and Sergey Edunov and Shaoliang Nie and Sharan Narang and Sharath Raparthy and Sheng Shen and Shengye Wan and Shruti Bhosale and Shun Zhang and Simon Vandenhende and Soumya Batra and Spencer Whitman and Sten Sootla and Stephane Collot and Suchin Gururangan and Sydney Borodinsky and Tamar Herman and Tara Fowler and Tarek Sheasha and Thomas Georgiou and Thomas Scialom and Tobias Speckbacher and Todor Mihaylov and Tong Xiao and Ujjwal Karn and Vedanuj Goswami and Vibhor Gupta and Vignesh Ramanathan and Viktor Kerkez and Vincent Gonguet and Virginie Do and Vish Vogeti and Vítor Albiero and Vladan Petrovic and Weiwei Chu and Wenhan Xiong and Wenyin Fu and Whitney Meers and Xavier Martinet and Xiaodong Wang and Xiaofang Wang and Xiaoqing Ellen Tan and Xide Xia and Xinfeng Xie and Xuchao Jia and Xuewei Wang and Yaelle Goldschlag and Yashesh Gaur and Yasmine Babaei and Yi Wen and Yiwen Song and Yuchen Zhang and Yue Li and Yuning Mao and Zacharie Delpierre Coudert and Zheng Yan and Zhengxing Chen and Zoe Papakipos and Aaditya Singh and Aayushi Srivastava and Abha Jain and Adam Kelsey and Adam Shajnfeld and Adithya Gangidi and Adolfo Victoria and Ahuva Goldstand and Ajay Menon and Ajay Sharma and Alex Boesenberg and Alexei Baevski and Allie Feinstein and Amanda Kallet and Amit Sangani and Amos Teo and Anam Yunus and Andrei Lupu and Andres Alvarado and Andrew Caples and Andrew Gu and Andrew Ho and Andrew Poulton and Andrew Ryan and Ankit Ramchandani and Annie Dong and Annie Franco and Anuj Goyal and Aparajita Saraf and Arkabandhu Chowdhury and Ashley Gabriel and Ashwin Bharambe and Assaf Eisenman and Azadeh Yazdan and Beau James and Ben Maurer and Benjamin Leonhardi and Bernie Huang and Beth Loyd and Beto De Paola and Bhargavi Paranjape and Bing Liu and Bo Wu and Boyu Ni and Braden Hancock and Bram Wasti and Brandon Spence and Brani Stojkovic and Brian Gamido and Britt Montalvo and Carl Parker and Carly Burton and Catalina Mejia and Ce Liu and Changhan Wang and Changkyu Kim and Chao Zhou and Chester Hu and Ching-Hsiang Chu and Chris Cai and Chris Tindal and Christoph Feichtenhofer and Cynthia Gao and Damon Civin and Dana Beaty and Daniel Kreymer and Daniel Li and David Adkins and David Xu and Davide Testuggine and Delia David and Devi Parikh and Diana Liskovich and Didem Foss and Dingkang Wang and Duc Le and Dustin Holland and Edward Dowling and Eissa Jamil and Elaine Montgomery and Eleonora Presani and Emily Hahn and Emily Wood and Eric-Tuan Le and Erik Brinkman and Esteban Arcaute and Evan Dunbar and Evan Smothers and Fei Sun and Felix Kreuk and Feng Tian and Filippos Kokkinos and Firat Ozgenel and Francesco Caggioni and Frank Kanayet and Frank Seide and Gabriela Medina Florez and Gabriella Schwarz and Gada Badeer and Georgia Swee and Gil Halpern and Grant Herman and Grigory Sizov and Guangyi and Zhang and Guna Lakshminarayanan and Hakan Inan and Hamid Shojanazeri and Han Zou and Hannah Wang and Hanwen Zha and Haroun Habeeb and Harrison Rudolph and Helen Suk and Henry Aspegren and Hunter Goldman and Hongyuan Zhan and Ibrahim Damlaj and Igor Molybog and Igor Tufanov and Ilias Leontiadis and Irina-Elena Veliche and Itai Gat and Jake Weissman and James Geboski and James Kohli and Janice Lam and Japhet Asher and Jean-Baptiste Gaya and Jeff Marcus and Jeff Tang and Jennifer Chan and Jenny Zhen and Jeremy Reizenstein and Jeremy Teboul and Jessica Zhong and Jian Jin and Jingyi Yang and Joe Cummings and Jon Carvill and Jon Shepard and Jonathan McPhie and Jonathan Torres and Josh Ginsburg and Junjie Wang and Kai Wu and Kam Hou U and Karan Saxena and Kartikay Khandelwal and Katayoun Zand and Kathy Matosich and Kaushik Veeraraghavan and Kelly Michelena and Keqian Li and Kiran Jagadeesh and Kun Huang and Kunal Chawla and Kyle Huang and Lailin Chen and Lakshya Garg and Lavender A and Leandro Silva and Lee Bell and Lei Zhang and Liangpeng Guo and Licheng Yu and Liron Moshkovich and Luca Wehrstedt and Madian Khabsa and Manav Avalani and Manish Bhatt and Martynas Mankus and Matan Hasson and Matthew Lennie and Matthias Reso and Maxim Groshev and Maxim Naumov and Maya Lathi and Meghan Keneally and Miao Liu and Michael L. Seltzer and Michal Valko and Michelle Restrepo and Mihir Patel and Mik Vyatskov and Mikayel Samvelyan and Mike Clark and Mike Macey and Mike Wang and Miquel Jubert Hermoso and Mo Metanat and Mohammad Rastegari and Munish Bansal and Nandhini Santhanam and Natascha Parks and Natasha White and Navyata Bawa and Nayan Singhal and Nick Egebo and Nicolas Usunier and Nikhil Mehta and Nikolay Pavlovich Laptev and Ning Dong and Norman Cheng and Oleg Chernoguz and Olivia Hart and Omkar Salpekar and Ozlem Kalinli and Parkin Kent and Parth Parekh and Paul Saab and Pavan Balaji and Pedro Rittner and Philip Bontrager and Pierre Roux and Piotr Dollar and Polina Zvyagina and Prashant Ratanchandani and Pritish Yuvraj and Qian Liang and Rachad Alao and Rachel Rodriguez and Rafi Ayub and Raghotham Murthy and Raghu Nayani and Rahul Mitra and Rangaprabhu Parthasarathy and Raymond Li and Rebekkah Hogan and Robin Battey and Rocky Wang and Russ Howes and Ruty Rinott and Sachin Mehta and Sachin Siby and Sai Jayesh Bondu and Samyak Datta and Sara Chugh and Sara Hunt and Sargun Dhillon and Sasha Sidorov and Satadru Pan and Saurabh Mahajan and Saurabh Verma and Seiji Yamamoto and Sharadh Ramaswamy and Shaun Lindsay and Shaun Lindsay and Sheng Feng and Shenghao Lin and Shengxin Cindy Zha and Shishir Patil and Shiva Shankar and Shuqiang Zhang and Shuqiang Zhang and Sinong Wang and Sneha Agarwal and Soji Sajuyigbe and Soumith Chintala and Stephanie Max and Stephen Chen and Steve Kehoe and Steve Satterfield and Sudarshan Govindaprasad and Sumit Gupta and Summer Deng and Sungmin Cho and Sunny Virk and Suraj Subramanian and Sy Choudhury and Sydney Goldman and Tal Remez and Tamar Glaser and Tamara Best and Thilo Koehler and Thomas Robinson and Tianhe Li and Tianjun Zhang and Tim Matthews and Timothy Chou and Tzook Shaked and Varun Vontimitta and Victoria Ajayi and Victoria Montanez and Vijai Mohan and Vinay Satish Kumar and Vishal Mangla and Vlad Ionescu and Vlad Poenaru and Vlad Tiberiu Mihailescu and Vladimir Ivanov and Wei Li and Wenchen Wang and Wenwen Jiang and Wes Bouaziz and Will Constable and Xiaocheng Tang and Xiaojian Wu and Xiaolan Wang and Xilun Wu and Xinbo Gao and Yaniv Kleinman and Yanjun Chen and Ye Hu and Ye Jia and Ye Qi and Yenda Li and Yilin Zhang and Ying Zhang and Yossi Adi and Youngjin Nam and Yu and Wang and Yu Zhao and Yuchen Hao and Yundi Qian and Yunlu Li and Yuzi He and Zach Rait and Zachary DeVito and Zef Rosnbrick and Zhaoduo Wen and Zhenyu Yang and Zhiwei Zhao and Zhiyu Ma},
      year={2024},
      eprint={2407.21783},
      archivePrefix={arXiv},
      primaryClass={cs.AI},
      url={https://arxiv.org/abs/2407.21783}, 
}

@misc{qwen25,
      title={Qwen2.5 Technical Report}, 
      author={An Yang and Baosong Yang and Beichen Zhang and Binyuan Hui and Bo Zheng and Bowen Yu and Chengyuan Li and Dayiheng Liu and Fei Huang and Haoran Wei and Huan Lin and Jian Yang and Jianhong Tu and Jianwei Zhang and Jianxin Yang and Jiaxi Yang and Jingren Zhou and Junyang Lin and Kai Dang and Keming Lu and Keqin Bao and Kexin Yang and Le Yu and Mei Li and Mingfeng Xue and Pei Zhang and Qin Zhu and Rui Men and Runji Lin and Tianhao Li and Tianyi Tang and Tingyu Xia and Xingzhang Ren and Xuancheng Ren and Yang Fan and Yang Su and Yichang Zhang and Yu Wan and Yuqiong Liu and Zeyu Cui and Zhenru Zhang and Zihan Qiu},
      year={2025},
      eprint={2412.15115},
      archivePrefix={arXiv},
      primaryClass={cs.CL},
      url={https://arxiv.org/abs/2412.15115}, 
}

@misc{mistral7b,
      title={Mistral 7B}, 
      author={Albert Q. Jiang and Alexandre Sablayrolles and Arthur Mensch and Chris Bamford and Devendra Singh Chaplot and Diego de las Casas and Florian Bressand and Gianna Lengyel and Guillaume Lample and Lucile Saulnier and Lélio Renard Lavaud and Marie-Anne Lachaux and Pierre Stock and Teven Le Scao and Thibaut Lavril and Thomas Wang and Timothée Lacroix and William El Sayed},
      year={2023},
      eprint={2310.06825},
      archivePrefix={arXiv},
      primaryClass={cs.CL},
      url={https://arxiv.org/abs/2310.06825}, 
}

@article{huberloss,
  title={Robust Estimation of a Location Parameter},
  author={Huber, Peter J},
  journal={The Annals of Mathematical Statistics},
  volume={35},
  number={1},
  pages={73--101},
  year={1964},
  doi={10.1214/aoms/1177703732},
  publisher={Institute of Mathematical Statistics}
}

@inproceedings{joshi2017triviaqa,
    title = "{T}rivia{QA}: A Large Scale Distantly Supervised Challenge Dataset for Reading Comprehension",
    author = "Joshi, Mandar  and
      Choi, Eunsol  and
      Weld, Daniel  and
      Zettlemoyer, Luke",
    booktitle = "Proceedings of the 55th Annual Meeting of the Association for Computational Linguistics (Volume 1: Long Papers)",
    month = jul,
    year = "2017",
    publisher = "Association for Computational Linguistics",
    url = "https://aclanthology.org/P17-1147/",
    doi = "10.18653/v1/P17-1147",
    pages = "1601--1611"
}

@inproceedings{
hendrycks2021measuring,
title={Measuring Massive Multitask Language Understanding},
author={Dan Hendrycks and Collin Burns and Steven Basart and Andy Zou and Mantas Mazeika and Dawn Song and Jacob Steinhardt},
booktitle={International Conference on Learning Representations},
year={2021},
url={https://openreview.net/forum?id=d7KBjmI3GmQ}
}

@inproceedings{chen2024inside,
title={{INSIDE}: {LLM}s' Internal States Retain the Power of Hallucination Detection},
author={Chao Chen and Kai Liu and Ze Chen and Yi Gu and Yue Wu and Mingyuan Tao and Zhihang Fu and Jieping Ye},
booktitle={The Twelfth International Conference on Learning Representations},
year={2024},
url={https://openreview.net/forum?id=Zj12nzlQbz}
}

@article{kwiatkowski2019natural,
    title = "Natural Questions: A Benchmark for Question Answering Research",
    author = "Kwiatkowski, Tom  and
      Palomaki, Jennimaria  and
      Redfield, Olivia  and
      Collins, Michael  and
      Parikh, Ankur  and
      Alberti, Chris  and
      Epstein, Danielle  and
      Polosukhin, Illia  and
      Devlin, Jacob  and
      Lee, Kenton  and
      Toutanova, Kristina  and
      Jones, Llion  and
      Kelcey, Matthew  and
      Chang, Ming-Wei  and
      Dai, Andrew M.  and
      Uszkoreit, Jakob  and
      Le, Quoc  and
      Petrov, Slav",
    journal = "Transactions of the Association for Computational Linguistics",
    volume = "7",
    year = "2019",
    publisher = "MIT Press",
    url = "https://aclanthology.org/Q19-1026/",
    doi = "10.1162/tacl_a_00276",
    pages = "452--466",
}

@inproceedings{li2023halueval,
    title = "{H}alu{E}val: A Large-Scale Hallucination Evaluation Benchmark for Large Language Models",
    author = "Li, Junyi  and
      Cheng, Xiaoxue  and
      Zhao, Xin  and
      Nie, Jian-Yun  and
      Wen, Ji-Rong",
    booktitle = "Proceedings of the 2023 Conference on Empirical Methods in Natural Language Processing",
    month = dec,
    year = "2023",
    publisher = "Association for Computational Linguistics",
    url = "https://aclanthology.org/2023.emnlp-main.397/",
    doi = "10.18653/v1/2023.emnlp-main.397",
    pages = "6449--6464"
}

@inproceedings{min2023factscore,
    title = "{FA}ct{S}core: Fine-grained Atomic Evaluation of Factual Precision in Long Form Text Generation",
    author = "Min, Sewon  and
      Krishna, Kalpesh  and
      Lyu, Xinxi  and
      Lewis, Mike  and
      Yih, Wen-tau  and
      Koh, Pang  and
      Iyyer, Mohit  and
      Zettlemoyer, Luke  and
      Hajishirzi, Hannaneh",
    booktitle = "Proceedings of the 2023 Conference on Empirical Methods in Natural Language Processing",
    month = dec,
    year = "2023",
    publisher = "Association for Computational Linguistics",
    url = "https://aclanthology.org/2023.emnlp-main.741/",
    doi = "10.18653/v1/2023.emnlp-main.741",
    pages = "12076--12100"
}

@inproceedings{wan2025fastfact,
    title = "{F}a{S}t{F}act: Faster, Stronger Long-Form Factuality Evaluations in {LLM}s",
    author = "Wan, Yingjia  and
      Tan, Haochen  and
      Zhu, Xiao  and
      Zhou, Xinyu  and
      Li, Zhiwei  and
      Lv, Qingsong  and
      Sun, Changxuan  and
      Zeng, Jiaqi  and
      Xu, Yi  and
      Lu, Jianqiao  and
      Liu, Yinhong  and
      Guo, Zhijiang",
    booktitle = "Findings of the Association for Computational Linguistics: EMNLP 2025",
    month = nov,
    year = "2025",
    publisher = "Association for Computational Linguistics",
    url = "https://aclanthology.org/2025.findings-emnlp.1295/",
    doi = "10.18653/v1/2025.findings-emnlp.1295",
    pages = "23814--23854",
    ISBN = "979-8-89176-335-7",
}

@inproceedings{wang2024factcheck,
    title = "Factcheck-Bench: Fine-Grained Evaluation Benchmark for Automatic Fact-checkers",
    author = "Wang, Yuxia  and
      Gangi Reddy, Revanth  and
      Mujahid, Zain Muhammad  and
      Arora, Arnav  and
      Rubashevskii, Aleksandr  and
      Geng, Jiahui  and
      Mohammed Afzal, Osama  and
      Pan, Liangming  and
      Borenstein, Nadav  and
      Pillai, Aditya  and
      Augenstein, Isabelle  and
      Gurevych, Iryna  and
      Nakov, Preslav",
    booktitle = "Findings of the Association for Computational Linguistics: EMNLP 2024",
    month = nov,
    year = "2024",
    publisher = "Association for Computational Linguistics",
    url = "https://aclanthology.org/2024.findings-emnlp.830/",
    doi = "10.18653/v1/2024.findings-emnlp.830",
    pages = "14199--14230",
}

@inproceedings{hu2025llm,
author = {Hu, Beizhe and Sheng, Qiang and Cao, Juan and Li, Yang and Wang, Danding},
title = {LLM-Generated Fake News Induces Truth Decay in News Ecosystem: A Case Study on Neural News Recommendation},
year = {2025},
isbn = {9798400715921},
publisher = {Association for Computing Machinery},
url = {https://doi.org/10.1145/3726302.3730027},
doi = {10.1145/3726302.3730027},
booktitle = {Proceedings of the 48th International ACM SIGIR Conference on Research and Development in Information Retrieval},
pages = {435–445}
}

@inproceedings{hu2024bad,
  title={Bad actor, good advisor: Exploring the role of large language models in fake news detection},
  author={Hu, Beizhe and Sheng, Qiang and Cao, Juan and Shi, Yuhui and Li, Yang and Wang, Danding and Qi, Peng},
  booktitle={Proceedings of the AAAI conference on artificial intelligence},
  volume={38},
  number={20},
  pages={22105--22113},
  doi={10.1609/aaai.v38i20.30214},
  year={2024}
}

@article{zhao2026agentic,
  title = {An agentic system for rare disease diagnosis with traceable reasoning},
  volume = {651},
  ISSN = {1476-4687},
  url = {http://dx.doi.org/10.1038/s41586-025-10097-9},
  DOI = {10.1038/s41586-025-10097-9},
  journal = {Nature},
  author = {Zhao,  Weike and Wu,  Chaoyi and Fan,  Yanjie and Qiu,  Pengcheng and Zhang,  Xiaoman and Sun,  Yuze and Zhou,  Xiao and Zhang,  Shuju and Peng,  Yu and Wang,  Yanfeng and Sun,  Xin and Zhang,  Ya and Yu,  Yongguo and Sun,  Kun and Xie,  Weidi},
  year = {2026},
  pages = {775–784}
}

@inproceedings{chen2026simulating,
  title={Simulating Dispute Mediation with LLM-Based Agents for Legal Research},
  author={Chen, Junjie and Li, Haitao and Qin, Minghao and Zhou, Yujia and Ren, Yanxue and Wang, Wuyue and Liu, Yiqun and Wu, Yueyue and Ai, Qingyao},
  booktitle={Proceedings of the AAAI Conference on Artificial Intelligence},
  volume={40},
  number={35},
  pages={29368--29375},
  year={2026},
  doi={10.1609/aaai.v40i35.40177}
}

@inproceedings{sun2025towards,
  title={Towards Detecting LLMs Hallucination via Markov Chain-based Multi-agent Debate Framework},
  author={Sun, Xiaoxi and Li, Jinpeng and Zhong, Yan and Zhao, Dongyan and Yan, Rui},
  booktitle={2025 IEEE International Conference on Acoustics, Speech and Signal Processing},
  pages={1--5},
  year={2025},
  publisher={IEEE},
  doi={10.1109/ICASSP49660.2025.10889448},
}

@inproceedings{chernfactool,
  title={FacTool: Factuality Detection in Generative AI--A Tool Augmented Framework for Multi-Task and Multi-Domain Scenarios},
  author={Chern, Ethan and Chern, Steffi and Chen, Shiqi and Yuan, Weizhe and Feng, Kehua and Zhou, Chunting and He, Junxian and Neubig, Graham and Liu, Pengfei},
  booktitle={Second Conference on Language Modeling},
  year={2025},
  url={https://openreview.net/forum?id=hJkQL9VtWT}
}

@inproceedings{Liu24tutorials,
author = {Liu, Aiwei and Sheng, Qiang and Hu, Xuming},
title = {Preventing and Detecting Misinformation Generated by Large Language Models},
year = {2024},
isbn = {9798400704314},
publisher = {Association for Computing Machinery},
url = {https://doi.org/10.1145/3626772.3661377},
doi = {10.1145/3626772.3661377},
booktitle = {Proceedings of the 47th International ACM SIGIR Conference on Research and Development in Information Retrieval},
pages = {3001–3004}
}

@inproceedings{
ni-etal-2025-towards,
title = "Towards Fully Exploiting {LLM} Internal States to Enhance Knowledge Boundary Perception",
author = "Ni, Shiyu  and
  Bi, Keping  and
  Guo, Jiafeng  and
  Yu, Lulu  and
  Bi, Baolong  and
  Cheng, Xueqi",
booktitle = "Proceedings of the 63rd Annual Meeting of the Association for Computational Linguistics (Volume 1: Long Papers)",
month = jul,
year = "2025",
publisher = "Association for Computational Linguistics",
url = "https://aclanthology.org/2025.acl-long.1184/",
doi = "10.18653/v1/2025.acl-long.1184",
pages = "24315--24329",
ISBN = "979-8-89176-251-0",
}

@inproceedings{
li2023inferencetime,
title={Inference-Time Intervention: Eliciting Truthful Answers from a Language Model},
author={Kenneth Li and Oam Patel and Fernanda Vi{\'e}gas and Hanspeter Pfister and Martin Wattenberg},
booktitle={Thirty-seventh Conference on Neural Information Processing Systems},
year={2023},
url={https://openreview.net/forum?id=aLLuYpn83y}
}

@misc{tan2026basecal,
      title={BaseCal: Unsupervised Confidence Calibration via Base Model Signals}, 
      author={Hexiang Tan and Wanli Yang and Junwei Zhang and Xin Chen and Rui Tang and Du Su and Jingang Wang and Yuanzhuo Wang and Fei Sun and Xueqi Cheng},
      year={2026},
      eprint={2601.03042},
      archivePrefix={arXiv},
      primaryClass={cs.CL},
      url={https://arxiv.org/abs/2601.03042}, 
}

@inproceedings{
li2025conftuner,
title={ConfTuner: Training Large Language Models to Express Their Confidence Verbally},
author={Yibo Li and Miao Xiong and Jiaying Wu and Bryan Hooi},
booktitle={The Thirty-ninth Annual Conference on Neural Information Processing Systems},
year={2025},
url={https://openreview.net/forum?id=VZQ04Ojhu5}
}

@misc{qi2026detectingcontextual,
      title={Detecting Contextual Hallucinations in LLMs with Frequency-Aware Attention}, 
      author={Siya Qi and Yudong Chen and Runcong Zhao and Qinglin Zhu and Zhanghao Hu and Wei Liu and Yulan He and Zheng Yuan and Lin Gui},
      year={2026},
      eprint={2602.18145},
      archivePrefix={arXiv},
      primaryClass={cs.CL},
      url={https://arxiv.org/abs/2602.18145}, 
}

@inproceedings{dhuliawala-etal-2024-chain,
    title = "Chain-of-Verification Reduces Hallucination in Large Language Models",
    author = "Dhuliawala, Shehzaad  and
      Komeili, Mojtaba  and
      Xu, Jing  and
      Raileanu, Roberta  and
      Li, Xian  and
      Celikyilmaz, Asli  and
      Weston, Jason",
    booktitle = "Findings of the Association for Computational Linguistics: ACL 2024",
    month = aug,
    year = "2024",
    publisher = "Association for Computational Linguistics",
    url = "https://aclanthology.org/2024.findings-acl.212/",
    doi = "10.18653/v1/2024.findings-acl.212",
    pages = "3563--3578",
}

@article{metaselfcorrecting,
author = {Zhang, Wei and Dai, Guojun and Luo, Ding and Wang, Yan and Ye, Chen},
title = {From Hallucination to Certainty: Meta-Knowledge Guided Self-Correcting Large Language Models},
year = {2026},
publisher = {Association for Computing Machinery},
issn = {2157-6904},
doi = {10.1145/3797906},
journal = {ACM Transactions on Intelligent Systems and Technology}
}

@inproceedings{preffend,
  title={Integrating pattern-and fact-based fake news detection via model preference learning},
  author={Sheng, Qiang and Zhang, Xueyao and Cao, Juan and Zhong, Lei},
  booktitle={Proceedings of the 30th ACM international conference on information \& knowledge management},
  pages={1640--1650},
  year={2021},
  doi={10.1145/3459637.3482440}
}

@inproceedings{yuanyige,
author = {Yuan, Yige and Xu, Bingbing and Tan, Hexiang and Sun, Fei and Xiao, Teng and Li, Wei and Shen, Huawei and Cheng, Xueqi},
title = {Fact-Level Calibration and Correction for Long-Form Generations},
year = {2025},
publisher = {Association for Computing Machinery},
url = {https://doi.org/10.1145/3726302.3730195},
doi = {10.1145/3726302.3730195},
booktitle = {Proceedings of the 48th International ACM SIGIR Conference on Research and Development in Information Retrieval},
pages = {2807–2811},
numpages = {5},
series = {SIGIR '25}
}

@inproceedings{prism,
    title = "Prompt-Guided Internal States for Hallucination Detection of Large Language Models",
    author = "Zhang, Fujie  and
      Yu, Peiqi  and
      Yi, Biao  and
      Zhang, Baolei  and
      Li, Tong  and
      Liu, Zheli",
    booktitle = "Proceedings of the 63rd Annual Meeting of the Association for Computational Linguistics (Volume 1: Long Papers)",
    month = jul,
    year = "2025",
    publisher = "Association for Computational Linguistics",
    url = "https://aclanthology.org/2025.acl-long.1058/",
    doi = "10.18653/v1/2025.acl-long.1058",
    pages = "21806--21818",
    ISBN = "979-8-89176-251-0",
}

@misc{bai2026inficheck,
      title={InFi-Check: Interpretable and Fine-Grained Fact-Checking of LLMs}, 
      author={Yuzhuo Bai and Shuzheng Si and Kangyang Luo and Qingyi Wang and Wenhao Li and Gang Chen and Fanchao Qi and Maosong Sun},
      year={2026},
      eprint={2601.06666},
      archivePrefix={arXiv},
      primaryClass={cs.CL},
      url={https://arxiv.org/abs/2601.06666}, 
}

@inproceedings{li-etal-2025-towards,
    title = "Towards Harmonized Uncertainty Estimation for Large Language Models",
    author = "Li, Rui  and
      Long, Jing  and
      Qi, Muge  and
      Xia, Heming  and
      Sha, Lei  and
      Wang, Peiyi  and
      Sui, Zhifang",
    booktitle = "Proceedings of the 63rd Annual Meeting of the Association for Computational Linguistics (Volume 1: Long Papers)",
    year = "2025",
    publisher = "Association for Computational Linguistics",
    url = "https://aclanthology.org/2025.acl-long.1118/",
    doi = "10.18653/v1/2025.acl-long.1118",
    pages = "22938--22953",
    ISBN = "979-8-89176-251-0",
}

\appendix

\vspace{-2mm}
\section{Additional Details of Datasets}
\subsection{Introduction of Used Datasets}

\label{app:dataset}

The detailed introduction of the four datasets used is as follows:
\begin{itemize}[leftmargin=*]
    \item \textbf{TriviaQA~\citep{joshi2017triviaqa}:} A large-scale dataset for reading comprehension and question answering, consisting of trivia questions authored by enthusiasts and their associated evidence documents. We utilize the deduplicated validation split \textit{(rc.nocontext subset)} with 9,961 Question-Answer Pairs.
    \vspace{1.5mm}
    \item \textbf{MMLU~\citep{hendrycks2021measuring}: } A benchmark covering 57 tasks across diverse domains, including STEM, humanities, social sciences, and professional knowledge. It evaluates a model’s multitasking and general reasoning abilities using multiple-choice questions drawn from publicly available exams and academic sources. We utilize the test set with 14,041 multiple-choice questions.
    \vspace{1.5mm}
    \item \textbf{NQ\_Open~\citep{kwiatkowski2019natural}:} An open-domain question answering benchmark derived from Natural Questions. It contains real user queries paired with relevant passages from Wikipedia, designed to evaluate models’ ability to retrieve and answer factual questions from large text corpora. We use its validation split without relevant passages, which contains 3,610 QA pairs.
    \vspace{1.5mm}
    \item \textbf{HaluEval~\citep{li2023halueval}:}  A large collection of generated and human-annotated hallucinated samples for evaluating the performance of LLMs in recognizing hallucination, including four subsets: QA, Dialogue, Summarization, and General Data. In our experiment, we use the QA split with 10K data samples.
\end{itemize}

\subsection{Prompts for Response and Self-judgment}
\label{app:prompt}
In this paper, we design the following simple prompt template to instruct the LLM to generate the response $O_r$:

\begin{tcolorbox}[
    title={Prompt for Response Generation}, 
    fonttitle=\bfseries, 
    halign title=left,           
    colback=white!95!gray, 
    colframe=black, 
    colbacktitle=black!10!white, 
    coltitle=black,              
    width=\linewidth, 
    rounded corners, 
    boxrule=0.8pt,
    lefttitle=2mm               
]
Answer the question concisely: \\
\textbf{Q:} \{$Q_r$\} \\
\textbf{A: }
\end{tcolorbox}

Then we design the following prompt for LLM self-judgment to get the verbal judgment $O_j$:

\begin{tcolorbox}[
    title={Prompt for LLM Self-Judgment}, 
    fonttitle=\bfseries, 
    halign title=left,          
    colback=white!95!gray, 
    colframe=black, 
    colbacktitle=black!10!white, 
    coltitle=black,              
    width=\linewidth, 
    rounded corners, 
    boxrule=0.8pt,
    lefttitle=2mm                
]
Given the following QA pair: \\
\textbf{Q:} \{$Q_r$\} \\
\textbf{A: }\{$O_r$\} \\
Does the answer above reflect the facts? \\
Please respond with one of the following labels: ``Yes'' or ``No''. \\
\textbf{Answer: }
\end{tcolorbox}

After generation, we filter out a small number of invalid responses (\eg, ‘…’), which fall outside the scope of hallucination detection. 

\subsection{Automatic Labeling of LLMs' Responses}
\label{app:label}

To obtain the factuality label $L_r$ for LLM responses, we design a three-stage automated annotation pipeline consisting of text pattern matching, semantic similarity scoring, and annotation using GPT-4o-mini. Each stage hierarchically processes the remaining unlabeled samples.

First, for samples that cannot be annotated via text pattern matching, we compute the semantic entailment score between the LLM response and the golden answer using the nli-deberta-v3-base model\footnote{\url{https://huggingface.co/cross-encoder/nli-deberta-v3-base}}
, which is specifically fine-tuned for natural language inference. Next, for samples with neutral NLI scores, we employ GPT-4o-mini as the annotator to assign factuality labels, using the prompt template shown below.

\begin{tcolorbox}[
    title={Prompt for GPT-4o-mini Annotation}, 
    fonttitle=\bfseries, 
    halign title=left,           
    colback=white!95!gray, 
    colframe=black, 
    colbacktitle=black!10!white, 
    coltitle=black,              
    width=\linewidth, 
    rounded corners, 
    boxrule=0.8pt,
    lefttitle=2mm                
]
Given the following Query--Response pair: \\
\textbf{Query:} \{$Q_r$\} \\
\textbf{Response:} \{$O_r$\} \\[2pt]
A possible ground-truth answer is also provided: \\
\textbf{Ground-Truth (possible):} \{$G_t$\} \\[4pt]
Your task is to determine whether the Response is accurate, based on: \\
1) The provided ground-truth (possible), \\
2) Your own knowledge. \\[4pt]
Please choose one of the following labels as your final judgment: \\
\texttt{``True''}, \texttt{``False''}, \texttt{``Uncertain''}. \\[4pt]
\textbf{Final Judgment:}
\end{tcolorbox}

For samples labeled as ``Uncertain'' by GPT-4o-mini, we conduct manual verification and selectively discard them due to the ambiguity and potential unreliability of their factuality labels. To assess annotation quality, we randomly sample 200 such uncertain instances for manual inspection, achieving an agreement rate of 93.50\%, which indicates that GPT-4o-mini provides sufficiently reliable annotations for effective data filtering.

Finally, to further check the overall annotation quality, we randomly sample 800 instances from the automatically labeled dataset for manual verification. The agreement between automatic and human annotations reaches 96.125\%, demonstrating the high quality of the automated labeling pipeline. The statistics of the final available dataset and label distribution are presented in Table~\ref{tab:dataset_statistics}.

\begin{table}[htbp]
\centering\small
\captionsetup{justification=centering}
\caption{Dataset statistics across benchmarks and large language models used in the experiments}
\setlength{\tabcolsep}{3.7pt}       

\begin{tabular}{l ccc ccc ccc ccc}
\toprule
\multirow{2}{*}[-0.5em]{\textbf{LLM}} 
& \multicolumn{3}{c}{\textbf{TriviaQA}} 
& \multicolumn{3}{c}{\textbf{MMLU}} 
& \multicolumn{3}{c}{\textbf{NQ\_Open}} 
& \multicolumn{3}{c}{\textbf{HaluEval}} \\
\cmidrule(lr){2-4} \cmidrule(lr){5-7} \cmidrule(lr){8-10} \cmidrule(lr){11-13}
& Total & Real & Hallu.
& Total & Real & Hallu.
& Total & Real & Hallu.
& Total & Real & Hallu. \\
\midrule
Llama-3.1-8B-Instruct
& 9,668 & 6,981 & 2,687
& 10,555 & 6,863 & 3,692
& 3,255 & 2,183 & 1,072
& 8,976 & 3,784 & 5,192 \\

Llama-3.1-70B-Instruct
& 9,848 & 8,485 & 1,363
& 9,112 & 7,518 & 1,594
& 3,400 & 2,638 & 762
& 9,438 & 5,225 & 4,213 \\

Qwen2.5-32B-Instruct
& 9,702 & 7,161 & 2,541
& 10,574 & 8,643 & 1,931
& 3,364 & 2,147 & 1,217
& 9,296 & 4,345 & 4,951 \\

Mistral-7B-Instruct-v0.3
& 9,752 & 6,859 & 2,893
& 10,750 & 6,612 & 4,138
& 3,316 & 2,072 & 1,244
& 9,203 & 3,970 & 5,233 \\
\bottomrule
\end{tabular}
\label{tab:dataset_statistics}
\end{table}

\section{Details of Intrinsic Patterns Extraction}
\label{app:feature}
\subsection{Hidden States ($H_r$ \& $H_j$)}
\paragraph{Selection Strategy of $K_{\text{val}}$.}
To balance performance and efficiency while avoiding the increasing complexity of searching for the optimal layer as LLM size grows, we adopt a quantile-based strategy. Specifically, we partition the total number of LLM layers according to proportional quantiles $[1/8, 1/4, 3/8, 1/2, 5/8, 3/4, 7/8, 1]$, yielding eight candidate layers. During the implementation of SAPLMA, we perform a grid search over the hidden states of these candidate layers, and select the one achieving the highest Macro-F1 score on the validation set as $K_{\text{val}, r}$ and $K_{\text{val}, j}$ for the response and self-judgment settings, respectively.

\subsection{Prediction Logits ($P_r$ \& $P_j$)}
\label{app:logits}
\paragraph{Additional Details of Logits Lens.}
Given a natural language input $X$, we first tokenize it into a sequence of tokens $T = [t_0, t_1, \ldots, t_{N-1}]$ using the tokenizer associated with the target LLM, where $N$ denotes the number of tokens. The token sequence is then mapped to embedding vectors $E = [e_0, e_1, \ldots, e_{N-1}]$ via the embedding matrix. Subsequently, $E$ is processed through a stack of Transformer blocks to produce hidden states $H = [h_n^k]$, where $h_n^k$ represents the hidden state of the $n$-th token at the $k$-th layer. The hidden state of the final token at the last layer, $h_{N-1}^K$, is projected through the unembedding matrix to obtain the probability distribution for next-token prediction.\footnote{For brevity, we omit standard operations such as Layer Normalization and Softmax.}

For Logits Lens, let $N_{Q_r}$ and $N_{O_r}$ denote the token lengths of the query $Q_r$ and response $O_r$, respectively. To compute the probabilities of tokens in $O_r$, we first extract the hidden states corresponding to the token span $T[N_{Q_r}-1 : N_{Q_r}+N_{O_r}-1]$ across all layers, resulting in a tensor of shape $(N_{O_r}, \text{layer\_num}, \text{hidden\_dim})$. We then project these hidden states into the LLM's output space using the unembedding matrix, yielding a probability tensor of shape $(N_{O_r}, \text{layer\_num}, \text{vocab\_size})$.Finally, we obtain the layer-wise probabilities for each token in $O_r$ by indexing this tensor with the corresponding token IDs, resulting in a matrix of shape $(N_{O_r}, \text{layer\_num})$.\footnote{\text{vocab\_size} denotes the vocabulary size of the LLM, and \textit{layer\_num} corresponds to $K$ defined above.} 

To obtain a sequence-level Logits prediction representation, we employ a single multi-head Transformer layer followed by mean pooling to aggregate token-level representations in $P_r$. Similar design choices have been explored in prior works such as those from~\citet{wang2023seqxgpt} and \citet{shi2024ten}. For $P_j$, we extract the Logits Lens representation of the first token in $O_j$ (typically ``Yes'' or ``No'', along with their aggregated synonyms), and further enhance the contrast using Eq.~(1). The resulting representation is then used as a sequence-level feature, as it sufficiently captures the core decision semantics of the judgment.

\begin{tcolorbox}[colback=gray!8!white, colframe=gray!60!black, arc=1.5mm, boxrule=0.6pt, left=2mm, right=2mm, top=2mm, bottom=2mm]
\noindent\textbf{The equivalents of \textit{``Yes''} include:}\\[0.5ex]
\small 
\texttt{[`Yes', `yes', `YES', `\_Yes', `\_yes', `\_YES', `Y', `y', `\_Y', `\_y', `True', `true', `TRUE', `\_True', `\_true', `\_TRUE', `Correct', `correct', `CORRECT', `\_Correct', `\_correct', `\_CORRECT']}
\tcblower 
\noindent\textbf{The equivalents of \textit{``No''} include:}\\[0.5ex]
\small
\texttt{[`No', `no', `NO', `\_No', `\_no', `\_NO', `N', `n', `\_N', `\_n', `False', `false', `FALSE', `\_False', `\_false', `\_FALSE', `Incorrect', `incorrect', `INCORRECT', `\_Incorrect', `\_incorrect', `\_INCORRECT']}
\end{tcolorbox}

\paragraph{Synonyms Group for ``Yes''/``No''.}
We adopt a verbalization strategy for LLM self-judgments to more accurately capture the model's decision uncertainty. Specifically, following common practices in prompt engineering, we treat tokens with similar semantics or prefixes as equivalents of ``Yes'' or ``No''. The synonym groups for ``Yes'' and ``No'' in Section~\ref{sec:judgment method} are constructed by intersecting the above token lists with the vocabulary of the target LLM.

\subsection{Attention Scores ($A_r$ \& $A_j$)}
\paragraph{Context Segmentation.}
Building upon the core idea of the original ``Lookback'' approach, we adapt the context segmentation strategy to align with the specific prompt templates utilized for the Response and Self-Judgment tasks in our work. For the response scenario ($r$), we partition the context into four distinct segments: system prompt, query $Q_r$, response trigger, and preceding tokens of $O_r$. The first three segments correspond directly to the three respective lines of the ``Prompt for Response Generation'' template detailed in Appendix~\ref{app:prompt}. Similarly, for the self-judgment scenario ($j$), we partition the context into six distinct segments: Framing, Query, Response, Eval\_Query, Format, and Trigger. These six segments correspond respectively to the six clauses of the ``Prompt for LLM Self-Judgment'' template, also detailed in Appendix~\ref{app:prompt}.

\begin{figure}[htbp]
\centering
\includegraphics[width=0.95\textwidth]{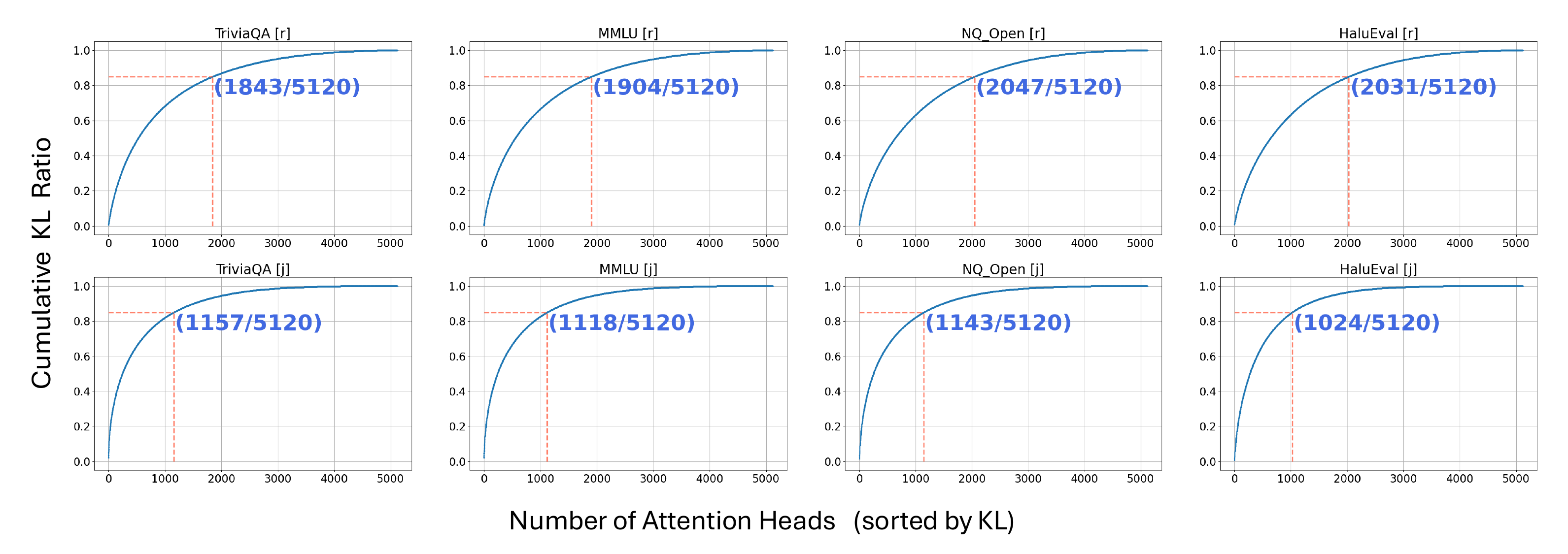}
\caption{Normalized cumulative distribution of KL divergence across all attention heads in Llama-3.1-70B. The pronounced long-tail distribution reveals that a small subset of heads captures the core factual discriminative capacity, while the majority provides only weak signals. \textcolor{orange}{The orange dashed line} indicates a cumulative probability of 0.85.}
\label{fig:attn}
\end{figure}

\paragraph{Lookback Ratio Calculation.}
We follow the core implementation by \citet{chuang2024lookback} and adapt it to our task setting.
Specifically, we first partition all tokens in a sequence into two categories: \emph{anchor tokens} and \emph{context tokens}. We then extract the token-level attention score matrix from anchor tokens to context tokens, with shape $(\text{layer\_num} \times \text{head\_num}, N_{\text{ach}}, N_{\text{ctx}})$, where $\text{head\_num}$ denotes the number of attention heads per layer in the LLM, and $N_{\text{ach}}$ and $N_{\text{ctx}}$ represent the number of anchor tokens and context tokens, respectively.
Next, according to the aforementioned context segmentation scheme, we divide the context tokens into $N_{\text{seg}}$ segments. For each anchor token, we perform intra-segment average pooling over the attention scores assigned to the context tokens within each segment, yielding \emph{ach2seg} attention distributions with shape $(\text{layer\_num} \times \text{head\_num}, N_{\text{ach}}, N_{\text{seg}})$.
We then normalize the \emph{ach2seg} attention map along the $N_{\text{seg}}$ dimension to obtain the token-level Lookback Ratio. Finally, we apply average pooling over all anchor tokens to derive the sequence-level Lookback Ratio, with shape $(\text{layer\_num} \times \text{head\_num}, N_{\text{seg}})$.
For the response scenario ($r$), the anchor is $O_r$, the context is $\text{concat}(Q_r, O_r)$, and $N_{\text{seg}} = 4$. For the self-judgment scenario ($j$), the anchor is $O_j$ (``Yes/No''), the context is $Q_j$, and $N_{\text{seg}} = 6$.

\paragraph{Top-$P$ Informative Heads Selection.}
As described above, the Lookback Ratio yields a high-dimensional vector of shape $(\text{layer\_num} \times \text{head\_num}, N_{\text{seg}})$ for each sequence. As LLMs scale, this dimensionality increases substantially (e.g., from 1024 heads in Llama-3.1-8B to 5120 in Llama-3.1-70B), raising the question of potential redundancy in these representations.

Specifically, we compute the KL divergence between the Lookback Ratio distributions of positive and negative training samples to quantify each head's factual discriminative power. We then rank the heads by descending KL divergence and plot the normalized cumulative distribution (as shown in Figure \ref{fig:attn}). The curve exhibits a sharp initial increase before flattening, indicating a pronounced long-tail distribution. This reveals that a small subset of heads captures the core factual discriminative capacity, while the majority provides only weak signals. Figure \ref{fig:attn} only shows results on Llama-3.1-70B for brevity, and consistent patterns hold across three other LLMs.

Such sparsity and redundancy limit effective training and introduce unnecessary computational overhead. Consequently, we apply a top-$P$ selection strategy ($P=0.85$, orange dashed line in Figure \ref{fig:attn}) to retain only heads with strong factual discriminability. Empirically, this approach preserves the hallucination detector's performance while effectively mitigating the long-tail effect and reducing the classifier's input dimensionality.

\section{Huber Loss Function}
\label{app:huber}
The Huber loss~\cite{huberloss} is a robust loss function that combines the advantages of Mean Squared Error (MSE) and Mean Absolute Error (MAE). Given a prediction $x$ and target $y$, the Huber loss is defined as:
\begin{equation}
\mathcal{L}_{\delta}(x, y) =
\begin{cases}
\frac{1}{2}(x - y)^2, & \text{if } |x - y| \leq \delta, \\
\delta \left(|x - y| - \frac{1}{2}\delta \right), & \text{otherwise},
\end{cases}
\end{equation}
where $\delta$ is a threshold hyperparameter that controls the transition between quadratic and linear regimes. In our framework, we set the threshold hyperparameter to $\delta = 0.5$.

Compared to MSE, which applies a quadratic penalty to all errors and is sensitive to outliers, the Huber loss behaves quadratically for small errors and linearly for large errors. This property makes it more robust to noisy or misaligned signals, while still maintaining smooth optimization near the optimum.

\section{Large Language Models}
\label{subsec:llms}
We use four commonly used open-source LLMs to cover different model families and scales.
For Llama 3~\cite{llama3}, we use Llama-3.1-8B-Instruct\footnote{\url{https://huggingface.co/meta-llama/Llama-3.1-8B-Instruct}} and Llama-3.1-70B-Instruct\footnote{\url{https://huggingface.co/meta-llama/Llama-3.1-70B-Instruct}}.
For Qwen-2.5~\cite{qwen25}, we use Qwen-2.5-32B-Instruct\footnote{\url{https://huggingface.co/Qwen/Qwen2.5-32B-Instruct}}.
For Mistral~\cite{mistral7b}, we use Mistral-7B-Instruct-v0.3\footnote{\url{https://huggingface.co/mistralai/Mistral-7B-Instruct-v0.3}}. The three LLM families are developed and released by independent organizations in different countries, and all of them are popular in the open-source community, which enhances their representativeness.

\section{Efficiency Analysis}

We conducted experiments on a server equipped with a single NVIDIA A800 GPU with a batch size of 128 and a learning rate of 1e-4 for the efficiency test.
As summarized in Table~\ref{tab:efficiency}, the \textit{Attns Based} variant demonstrates the highest training speed, requiring only $0.40$ s/epoch.
In terms of inference, the \textit{Hidden Based} variant achieves the lowest latency at $0.0215$ ms/instance.
Notably, although the \textit{Logits Based} variant has the fewest trainable parameters, it incurs higher temporal costs in both training ($1.63$ s/epoch) and testing phases compared to the other variants.
This indicates that our proposed LaaB framework would not introduce a significant increase in inference cost, making it a practical option for training hallucination detectors.

\begin{table}[htbp]
    \centering
    \small
    \caption{Efficiency profiling of three versions of LaaB. Parameters are reported in thousands (\textbf{K}). Inference latency is measured in \textbf{ms/instance}.}
    \label{tab:efficiency}
    \setlength{\tabcolsep}{2.5pt}
    \begin{tabular}{l c c c}
        \toprule
        \multirow{2}{*}{\textbf{Version}} & \textbf{Params (R/E)} & \textbf{Train Speed} & \textbf{Inf. Latency} \\
                                         & ($\times 10^3$)       & (s / epoch)          & (ms / inst.) \\
        \midrule
        Hidden Based & $1,090$ / $1,090$ & 1.06 & 0.0215 \\
        Logits Based & $155$ / $25$      & 1.63 & 0.0347 \\
        Attns Based  & $575$ / $520$     & 0.40 & 0.0232 \\
        \bottomrule
    \end{tabular}
\end{table}

\section{Introduction of Consistency-based Baselines}
\label{app:baselines}
Here, we supplement the introduction of two hallucination detection baseline methods based on multi-sampling consistency:
\begin{itemize}[leftmargin=*]
    \item \textbf{SelfCheckGPT~\citep{manakul2023selfcheckgpt}:} A sampling-based hallucination detector based on the assumption that the sampled responses will be consistent if an LLM has clear ``knowledge'' to respond to a query. If the LLM is hallucinating, the responses obtained from multiple samplings would be inconsistent. SelfCheckGPT provides five implementations, including BERTScore, MGQA, Unigram, NLI, and GPT-Prompt. In this work, we adopt SelfCheckGPT-NLI to balance detection performance and efficiency.
    \vspace{1.5mm}
    \item \textbf{Eigen-Score~\citep{chen2024inside}:} An uncertainty-based method to detect hallucinations by leveraging the semantic consistency of generated outputs in the dense embedding space. Specifically, it constructs a covariance matrix from the hidden states of multi-sampled responses and calculates its logarithmic determinant (equivalent to the sum of the logarithms of all eigenvalues) to measure the differential entropy of the representations. Therefore, a higher Eigen-Score indicates greater semantic divergence and a higher likelihood of hallucination. Following the original setup, we extract the hidden state of the final token from a middle layer for each sequence and report results without feature clipping in our experimental tables.
    
\end{itemize}

\section{Evaluation on More Baseline Methods}
To further demonstrate the broad adaptability of the LaaB framework across diverse hallucination detection approaches, we augment our experiments with two additional baselines: LapEigvals and TSV. Specifically, we compare the detection performance of the base detectors against their LaaB-enhanced counterparts. A brief introduction to these baselines is provided below:
\begin{itemize}[leftmargin=*]
    \item \textbf{LapEigvals~\citep{binkowski2025hallucination}:} An attention-based method that leverages the spectral properties of LLM attention maps. Specifically, it interprets the attention maps generated during the autoregressive inference process as weighted adjacency matrices of directed graphs. For each attention head across all transformer layers, the method computes the corresponding graph Laplacian matrix and extracts its top-$k$ eigenvalues. These eigenvalues serve to quantify disruptions or bottlenecks in the model's internal information flow, which are hypothesized to correlate with the occurrence of hallucinations. Ultimately, the extracted spectral features are concatenated, projected to a lower-dimensional space using Principal Component Analysis (PCA), and fed into a supervised probe to predict the final hallucination label.
    \vspace{1.5mm}
    \item \textbf{TSV~\citep{park2025steer}:} A hidden-based intervention method that reshapes the latent space of LLMs for hallucination detection. Pre-trained embeddings are optimized for linguistic coherence rather than factual accuracy, causing truthful and hallucinated representations to overlap. To mitigate this, TSV injects a single trainable steering vector into the residual stream of an intermediate transformer layer, scaled by a fixed strength hyperparameter. This lightweight intervention propagates through subsequent layers via inherent non-linear transformations, avoiding full model fine-tuning. As a result, the final-layer representations—often characterized by a von Mises-Fisher distribution—are reorganized into more compact and separable clusters, improving the linear separability between truthful and hallucinated outputs without degrading core language capabilities.

\end{itemize}

\vspace{10mm}
We follow the core designs of the two baseline methods and make light adaptations to align them with the LaaB framework. The detailed configurations are as follows:

\vspace{-3.5mm}
\paragraph{\underline{LapEigvals.}} We extract the top-10 Laplacian eigenvalues for each attention head and reduce the concatenated spectral features to 512 dimensions via PCA. Both the response detector $D_r$ and the self-judgment detector $D_j$ use an MLP probe with hidden sizes $[128, 32]$.

\vspace{-3.5mm}
\paragraph{\underline{TSV.}} The steering vector is injected into the Transformer residual stream with a strength of 5 and an EMA decay rate of 0.9. The intervention layer is chosen by model depth: layer 8 for Llama-3.1-8B-Instruct and layer 16 for Qwen2.5-32B-Instruct (approximately one-quarter depth), and layer 40 for Llama-3.1-70B-Instruct and layer 16 for Mistral-7B-Instruct-v0.3 (approximately halfway).
We randomly sample 2,000 training instances to learn separate steering vectors for the response and self-judgment scenarios. The LaaB constraint is then applied to the final-layer hidden representations after steering. Both $D_r$ and $D_j$ employ an MLP probe with hidden dimensions $[256, 128, 64]$. 
\vspace{1.5mm}

\begin{table}[htbp]
  \centering\small
   \caption{Performance comparison of LaaB with additional baseline methods (LapEigvals and TSV) in hallucination detection. The \textbf{blue-shaded} rows indicate the LaaB-enhanced versions. \textbf{Bolded} numbers denote that the use of LaaB outperforms its corresponding base version.}
   \setlength{\tabcolsep}{3.8pt} 
    \begin{tabular}{
      >{\centering\arraybackslash}m{2.2cm} 
      >{\raggedright\arraybackslash}m{2.0cm} 
      @{}c@{} 
      cc      
      >{\centering\arraybackslash}m{0.0cm} 
      cc      
      >{\centering\arraybackslash}m{0.0cm} 
      cc      
      >{\centering\arraybackslash}m{0.0cm} 
      cc      
      >{\centering\arraybackslash}m{0.0cm} 
      cc      
    }
    \toprule
     &  & \textbf{} & \multicolumn{2}{c}{\textbf{TriviaQA}} & \textbf{} & \multicolumn{2}{c}{\textbf{MMLU}} & \textbf{} & \multicolumn{2}{c}{\textbf{NQ\_Open}} & \textbf{} & \multicolumn{2}{c}{\textbf{HaluEval}} & \textbf{} & \multicolumn{2}{c}{\textbf{Average}} \\ \cline{4-5} \cline{7-8} \cline{10-11} \cline{13-14} \cline{16-17}
    \multirow{-2}{*}{\textbf{LLM}} & \multirow{-2}{*}{\textbf{Method}} &  & macF1 & Acc &  & macF1 & Acc &  & macF1 & Acc &  & macF1 & Acc & & macF1 & Acc \\ \midrule

     & LapEigvals &  & 74.92 & 78.32 &  & 68.77 & 69.75 &  & 69.56 & 72.71 &  & 75.81 & 76.04 & & 72.27 & 74.21 \\
     & {\hspace{10pt}\textit{+LaaB}} & \cellcolor[HTML]{ECF4FF} & \cellcolor[HTML]{ECF4FF}\textbf{76.44} & \cellcolor[HTML]{ECF4FF}\textbf{80.69} & \cellcolor[HTML]{ECF4FF} & \cellcolor[HTML]{ECF4FF}\textbf{69.71} & \cellcolor[HTML]{ECF4FF}\textbf{72.07} & \cellcolor[HTML]{ECF4FF} & \cellcolor[HTML]{ECF4FF}\textbf{70.48} & \cellcolor[HTML]{ECF4FF}\textbf{75.15} & \cellcolor[HTML]{ECF4FF} & \cellcolor[HTML]{ECF4FF}\textbf{76.45} & \cellcolor[HTML]{ECF4FF}\textbf{76.88} & \cellcolor[HTML]{ECF4FF} & \cellcolor[HTML]{ECF4FF}\textbf{73.27} & \cellcolor[HTML]{ECF4FF}\textbf{76.20} \\
     & TSV &  & 75.61 & 79.71 &  & 61.49 & 65.97 &  & 71.05 & 73.02 &  & 74.48 & 75.15 & & 70.66 & 73.46 \\
    \multirow{-4}{*}{\makecell[l]{\textbf{Llama-3.1-}\\\textbf{8B-Instruct}}} & {\hspace{10pt}\textit{+LaaB}} & \cellcolor[HTML]{ECF4FF} & \cellcolor[HTML]{ECF4FF}\textbf{77.17} & \cellcolor[HTML]{ECF4FF}\textbf{81.47} & \cellcolor[HTML]{ECF4FF} & \cellcolor[HTML]{ECF4FF}\textbf{62.59} & \cellcolor[HTML]{ECF4FF}\textbf{66.92} & \cellcolor[HTML]{ECF4FF} & \cellcolor[HTML]{ECF4FF}\textbf{71.49} & \cellcolor[HTML]{ECF4FF}\textbf{74.84} & \cellcolor[HTML]{ECF4FF} & \cellcolor[HTML]{ECF4FF}\textbf{74.82} & \cellcolor[HTML]{ECF4FF}\textbf{75.32} & \cellcolor[HTML]{ECF4FF} & \cellcolor[HTML]{ECF4FF}\textbf{71.52} & \cellcolor[HTML]{ECF4FF}\textbf{74.64} \\ \midrule

     & LapEigvals &  & 72.34 & 83.08 &  & 68.59 & 76.16 &  & 66.03 & 71.30 &  & 76.17 & 76.37 & & 70.78 & 76.73 \\
     & {\hspace{10pt}\textit{+LaaB}} & \cellcolor[HTML]{ECF4FF} & \cellcolor[HTML]{ECF4FF}\textbf{73.01} & \cellcolor[HTML]{ECF4FF}\textbf{86.37} & \cellcolor[HTML]{ECF4FF} & \cellcolor[HTML]{ECF4FF}\textbf{71.08} & \cellcolor[HTML]{ECF4FF}\textbf{81.97} & \cellcolor[HTML]{ECF4FF} & \cellcolor[HTML]{ECF4FF}\textbf{68.44} & \cellcolor[HTML]{ECF4FF}\textbf{77.01} & \cellcolor[HTML]{ECF4FF} & \cellcolor[HTML]{ECF4FF}\textbf{76.38} & \cellcolor[HTML]{ECF4FF}\textbf{76.48} & \cellcolor[HTML]{ECF4FF} & \cellcolor[HTML]{ECF4FF}\textbf{72.23} & \cellcolor[HTML]{ECF4FF}\textbf{80.46} \\
     & TSV &  & 71.85 & 85.41 &  & 71.12 & 81.48 &  & 66.95 & 74.52 &  & 77.85 & 77.96 & & 71.94 & 79.84 \\
    \multirow{-4}{*}{\makecell[l]{\textbf{Llama-3.1-}\\\textbf{70B-Instruct}}} & {\hspace{10pt}\textit{+LaaB}} & \cellcolor[HTML]{ECF4FF} & \cellcolor[HTML]{ECF4FF}\textbf{72.75} & \cellcolor[HTML]{ECF4FF}\textbf{86.47} & \cellcolor[HTML]{ECF4FF} & \cellcolor[HTML]{ECF4FF}71.10 & \cellcolor[HTML]{ECF4FF}\textbf{81.75} & \cellcolor[HTML]{ECF4FF} & \cellcolor[HTML]{ECF4FF}\textbf{68.17} & \cellcolor[HTML]{ECF4FF}\textbf{77.01} & \cellcolor[HTML]{ECF4FF} & \cellcolor[HTML]{ECF4FF}\textbf{78.09} & \cellcolor[HTML]{ECF4FF}\textbf{78.17} & \cellcolor[HTML]{ECF4FF} & \cellcolor[HTML]{ECF4FF}\textbf{72.53} & \cellcolor[HTML]{ECF4FF}\textbf{80.85} \\ \midrule

     & LapEigvals &  & 74.59 & 78.55 &  & 64.15 & 72.29 &  & 76.31 & 77.55 &  & 76.40 & 76.49 & & 72.86 & 76.22 \\
     & {\hspace{10pt}\textit{+LaaB}} & \cellcolor[HTML]{ECF4FF} & \cellcolor[HTML]{ECF4FF}\textbf{76.54} & \cellcolor[HTML]{ECF4FF}\textbf{81.53} & \cellcolor[HTML]{ECF4FF} & \cellcolor[HTML]{ECF4FF}\textbf{65.86} & \cellcolor[HTML]{ECF4FF}\textbf{77.15} & \cellcolor[HTML]{ECF4FF} & \cellcolor[HTML]{ECF4FF}\textbf{78.82} & \cellcolor[HTML]{ECF4FF}\textbf{80.06} & \cellcolor[HTML]{ECF4FF} & \cellcolor[HTML]{ECF4FF}\textbf{77.01} & \cellcolor[HTML]{ECF4FF}\textbf{77.08} & \cellcolor[HTML]{ECF4FF} & \cellcolor[HTML]{ECF4FF}\textbf{74.56} & \cellcolor[HTML]{ECF4FF}\textbf{78.96} \\
     & TSV &  & 76.21 & 80.86 &  & 63.27 & 69.18 &  & 76.47 & 77.55 &  & 77.20 & 77.29 & & 73.29 & 76.22 \\
    \multirow{-4}{*}{\makecell[l]{\textbf{Qwen-2.5-}\\\textbf{32B-Instruct}}} & {\hspace{10pt}\textit{+LaaB}} & \cellcolor[HTML]{ECF4FF} & \cellcolor[HTML]{ECF4FF}\textbf{78.12} & \cellcolor[HTML]{ECF4FF}\textbf{83.18} & \cellcolor[HTML]{ECF4FF} & \cellcolor[HTML]{ECF4FF}\textbf{64.21} & \cellcolor[HTML]{ECF4FF}\textbf{71.19} & \cellcolor[HTML]{ECF4FF} & \cellcolor[HTML]{ECF4FF}\textbf{78.30} & \cellcolor[HTML]{ECF4FF}\textbf{79.32} & \cellcolor[HTML]{ECF4FF} & \cellcolor[HTML]{ECF4FF}\textbf{77.82} & \cellcolor[HTML]{ECF4FF}\textbf{77.94} & \cellcolor[HTML]{ECF4FF} & \cellcolor[HTML]{ECF4FF}\textbf{74.61} & \cellcolor[HTML]{ECF4FF}\textbf{77.91} \\ \midrule

     & LapEigvals &  & 74.18 & 76.66 &  & 68.27 & 68.99 &  & 73.57 & 75.11 &  & 73.41 & 73.75 & & 72.36 & 73.63 \\
     & {\hspace{10pt}\textit{+LaaB}} & \cellcolor[HTML]{ECF4FF} & \cellcolor[HTML]{ECF4FF}\textbf{74.91} & \cellcolor[HTML]{ECF4FF}\textbf{78.05} & \cellcolor[HTML]{ECF4FF} & \cellcolor[HTML]{ECF4FF}\textbf{68.31} & \cellcolor[HTML]{ECF4FF}68.99 & \cellcolor[HTML]{ECF4FF} & \cellcolor[HTML]{ECF4FF}\textbf{74.59} & \cellcolor[HTML]{ECF4FF}\textbf{76.16} & \cellcolor[HTML]{ECF4FF} & \cellcolor[HTML]{ECF4FF}\textbf{74.08} & \cellcolor[HTML]{ECF4FF}\textbf{74.67} & \cellcolor[HTML]{ECF4FF} & \cellcolor[HTML]{ECF4FF}\textbf{72.97} & \cellcolor[HTML]{ECF4FF}\textbf{74.47} \\
     & TSV &  & 76.69 & 79.53 &  & 68.16 & 69.96 &  & 75.90 & 76.76 &  & 76.20 & 76.46 & & 74.24 & 75.68 \\
    \multirow{-4}{*}{\makecell[l]{\textbf{Mistral-7B-}\\\textbf{Instruct-v0.3}}} & {\hspace{10pt}\textit{+LaaB}} & \cellcolor[HTML]{ECF4FF} & \cellcolor[HTML]{ECF4FF}76.60 & \cellcolor[HTML]{ECF4FF}\textbf{79.58} & \cellcolor[HTML]{ECF4FF} & \cellcolor[HTML]{ECF4FF}\textbf{69.14} & \cellcolor[HTML]{ECF4FF}\textbf{70.24} & \cellcolor[HTML]{ECF4FF} & \cellcolor[HTML]{ECF4FF}\textbf{76.50} & \cellcolor[HTML]{ECF4FF}\textbf{77.66} & \cellcolor[HTML]{ECF4FF} & \cellcolor[HTML]{ECF4FF}74.44 & \cellcolor[HTML]{ECF4FF}75.54 & \cellcolor[HTML]{ECF4FF} & \cellcolor[HTML]{ECF4FF}74.17 & \cellcolor[HTML]{ECF4FF}\textbf{75.76} \\ \bottomrule
    \end{tabular}
    \label{tab:appendix-baselines}
\end{table}
\vspace{1mm}
The results in Table~\ref{tab:appendix-baselines} show that the LaaB framework improves the performance of both LapEigvals and TSV in most cases. This further demonstrates that, by introducing logical consistency constraints from both the response and self-judgment perspectives, LaaB can be reliably integrated with diverse hallucination detectors, highlighting its effectiveness and compatibility.

\section{Algorithm of LaaB}
\label{app:algorithm}
The pseudocode for the training and inference procedures of the LaaB hallucination detection framework is shown in Algorithm \ref{alg:laab}.

\begin{algorithm*}[t]
\caption{Training \& Inference of LaaB}
\label{alg:laab}
\small
\begin{algorithmic}
\Require Training set $\mathcal{S}$, Validation set $\mathcal{S}_{val}$, Testing set $\mathcal{S}_{test}$, Learning rates $(\eta_r, \eta_j, \eta_{tune})$, Early stopping patience $P$.
\Ensure Response detector $D_r(\theta_r)$, Self-Judgment detector $D_j(\theta_j)$.
\Procedure{\textbf{Training}}{$\mathcal{S}, \mathcal{S}_{val}$}
    \vspace{2pt}
    \Statex \textbf{\textsc{Stage 1: Asynchronous Mutual Learning}} \hfill \textit{/* $D_r$ and $D_j$ learn interactively */}
    \State $stop_r \gets \text{False}$, $stop_j \gets \text{False}$
    \While{\textbf{not} ($stop_r$ \textbf{and} $stop_j$)}
        \For{mini-batch $B \in \mathcal{S}$}
            \State \textbf{1.} Extract intrinsic features $F_r, F_j$ for $B$
            
            \If{\textbf{not} $stop_r$}
                \State \textbf{2.} $\mathcal{L}_{\text{CE}, r}, \mathcal{L}_{\text{Logic},r} \gets \text{MutualLoss}(D_r, D_j, F_r, F_j, S, role=r)$
                \State \textbf{3.} $\alpha_r \gets \text{AdaptiveWeight}(\mathcal{L}_{\text{CE}, r}, \mathcal{L}_{\text{Logic},r}, \theta_r)$
                \State \textbf{4.} $\mathcal{L}_r \gets \mathcal{L}_{\text{CE}, r} + \alpha_r \mathcal{L}_{\text{Logic}, r}$
                \State \textbf{5.} $\theta_r \gets \theta_r - \eta_r \nabla_{\theta_r} \mathcal{L}_r$
            \EndIf
            
            \If{\textbf{not} $stop_j$}
                \State \textbf{6.} $\mathcal{L}_{\text{CE}, j}, \mathcal{L}_{\text{Logic},j} \gets \text{MutualLoss}(D_r, D_j, F_r, F_j, S, role=j)$
                \State \textbf{7.} $\alpha_j \gets \text{AdaptiveWeight}(\mathcal{L}_{\text{CE}, j}, \mathcal{L}_{\text{Logic},j}, \theta_j)$
                \State \textbf{8.} $\mathcal{L}_j \gets \mathcal{L}_{\text{CE}, j} + \alpha_j \mathcal{L}_{\text{Logic}, j}$
                \State \textbf{9.} $\theta_j \gets \theta_j - \eta_j \nabla_{\theta_j} \mathcal{L}_j$
            \EndIf
        \EndFor
        
        \vspace{2pt}
        \Statex \hspace{\algorithmicindent} \textit{// Early Stopping \& State Reversion}
        \State Evaluate BCE loss on $\mathcal{S}_{val}$
        \State \textbf{If} $D_r$ ($D_j$) does not improve for $P$ epochs, freeze $\theta_r$ ($\theta_j$) and set $stop_r$ ($stop_j$) $\gets \text{True}$
        \State \textbf{If} one converges, revert to its best state and continue training the other
    \EndWhile
    
    \vspace{4pt}
    \Statex \textbf{\textsc{Stage 2: Joint Fine-tuning}} \hfill \textit{/* Synchronized optimization */}
    \State Restore $D_r$ and $D_j$ to their best states from Stage 1 and unfreeze
    \State $wait \gets 0$, $stop \gets \text{False}$
    \While{\textbf{not} $stop$}
        \For{mini-batch $B \in \mathcal{S}$}
            \State \textbf{1.} $\mathcal{L}_{\text{CE}, r},\mathcal{L}_{\text{CE}, j}, \mathcal{L}_{\text{Logic}} = \text{JointLoss}(D_r,D_j,F_r,F_j,S)$
            \State \textbf{2.} $\alpha \gets \frac{1}{2} \sum_{k \in \{r, j\}} \text{AdaptiveWeight}(\mathcal{L}_{\text{CE}, k}, \mathcal{L}_{\text{Logic}}, \theta_k)$
            \State \textbf{3.} $\mathcal{L}_{\text{Joint}} \gets \mathcal{L}_{\text{CE}, r} + \mathcal{L}_{\text{CE}, j} + \alpha \mathcal{L}_{\text{Logic}}$
            \State \textbf{4.} $\theta_r \gets \theta_r - \eta_{tune} \nabla_{\theta_r} \mathcal{L}_{\text{Joint}}$, $\theta_j \gets \theta_j - \eta_{tune} \nabla_{\theta_j} \mathcal{L}_{\text{Joint}}$
        \EndFor
        
        \vspace{2pt}
        \State Evaluate $\mathcal{L}_{\text{CE}, r}$ on $\mathcal{S}_{val}$: update best $\theta_r^*, \theta_j^*$ if improved, else $wait \gets wait + 1$
        \State \textbf{If} $wait \ge P$ \textbf{then break} \hfill \textit{/* Early stopping */}
    \EndWhile
    \State \Return Optimal parameters $\theta_r^*, \theta_j^*$
\EndProcedure

\Statex \hrulefill

\Procedure{\textbf{Inference}}{$\{Q_r, O_r\} \in S_{test}$}
    \Statex \textbf{\textsc{Deployment}} \hfill \Comment{Only $D_r$ is deployed to save costs}
    \State \textbf{1.} Extract intrinsic feature $F_r \in \{H_r, P_r, A_r\}$ during generating $O_r$
    \State \textbf{2.} Predict hallucination probability $S_r = D_r^*(F_r)$ 
    \State \Return $S_r$
\EndProcedure
\end{algorithmic}
\end{algorithm*}

\end{document}